%% file: main_arxiv.tex
\documentclass[11pt]{article}
\usepackage[utf8]{inputenc}

\usepackage{natbib}

\usepackage[utf8]{inputenc} 
\usepackage[T1]{fontenc}    
\usepackage{url}            
\usepackage{booktabs, tabularx}       
\usepackage{amsfonts}       
\usepackage{nicefrac}       
\usepackage[USenglish,british,american,australian,english]{babel}
\usepackage{amsthm,amsmath,amssymb}
\usepackage{multirow}
\usepackage[pdftex]{graphicx}
\usepackage{caption}
\usepackage{array}
\usepackage{enumitem} 
\usepackage{fullpage}
\usepackage[algo2e]{algorithm2e}
\usepackage{bm}
\usepackage{bbm}
\usepackage{smile_new}
\usepackage{lipsum}
\usepackage{mathrsfs}
\usepackage{dsfont}
\usepackage{epstopdf}

\usepackage{listings}
\usepackage[usenames,dvipsnames,svgnames,table]{xcolor}
\usepackage[colorlinks=true,
            linkcolor=red,
            urlcolor=blue,
            citecolor=blue]{hyperref}

\usepackage[activate={true,nocompatibility},final,tracking=true,kerning=true,spacing=true,factor=2000,stretch=10,shrink=10]{microtype}

\def\##1\#{\begin{align}#1\end{align}}
\def\$#1\${\begin{align*}#1\end{align*}}

\begin{document}

\title {\huge More Supervision, Less Computation: Statistical-Computational Tradeoffs in Weakly Supervised Learning\thanks{This work has been published in the Thirtieth Annual Conference on Neural Information Processing Systems (NeurIPS 2016).}}

\author{ 
Xinyang Yi$^\natural$\thanks{Google Brain}\quad~Zhaoran Wang$^\natural$\thanks{Northwestern University}\quad~ 
Zhuoran Yang$^\natural$\thanks{Princeton University}\quad~ Constantine Caramanis\thanks{University of Texas at Austin}\quad~Han Liu\thanks{Northwestern University} \\
\texttt{\small\{$\natural$: equal contribution\}}
}

\date{}

\maketitle

\begin{abstract}
We consider the weakly supervised binary classification problem where the labels are randomly flipped with probability $1-\alpha$. Although there exist numerous algorithms for this problem, it remains theoretically unexplored how the statistical accuracies and computational efficiency of these algorithms depend on the degree of supervision, which is quantified by $\alpha$. In this paper, we characterize the effect of $\alpha$ by establishing the information-theoretic and computational boundaries, namely, the minimax-optimal statistical accuracy that can be achieved by all algorithms, and polynomial-time algorithms under an oracle computational model. For small $\alpha$, our result shows a gap between these two boundaries, which represents the computational price of achieving the information-theoretic boundary due to the lack of supervision. Interestingly, we also show that this gap narrows as $\alpha$ increases. In other words, having more supervision, i.e., more correct labels, not only improves the optimal statistical accuracy as expected, but also enhances the computational efficiency for achieving such accuracy.
\end{abstract}

\input{intro_nips}

\input{background_nips}
\input{main_results_nips}

\section*{Acknowledgments}
We would like to thank   Vitaly Feldman for valuable discussions.
  
\bibliographystyle{ims}
 
\bibliography{graphbib}

\clearpage
\newpage
\input{proof}

\end{document}

%% file: intro_nips.tex

 \section{Introduction}\label{sec:intro}
Practical classification problems usually involve corrupted labels. Specifically, let $\{(\xb_i, z_i)\}_{i=1}^n$ be $n$ independent data points, where $\xb_i \in \RR^{d}$ is the covariate vector and $z_i \in \{0,1\}$ is the uncorrupted label. Instead of observing $\{(\xb_i, z_i)\}_{i=1}^n$, we observe $\{(\xb_i, y_i)\}_{i=1}^n$ in which $y_i$ is the corrupted label. In detail, with probability $(1-\alpha)$, $y_i$ is chosen uniformly at random over $\{0,1\}$, and with probability $\alpha$, $y_i = z_i$. Here $\alpha \in [0,1]$ quantifies the degree of supervision: a larger $\alpha$ indicates more supervision since we have more uncorrupted labels in this case. In this paper, we are particularly interested in the effect of $\alpha$ on the statistical accuracy and computational efficiency for parameter estimation in this problem, particularly in the high dimensional settings where the dimension $d$ is much larger than the sample size $n$.

There exists a vast body of literature on binary classification problems with corrupted labels. In particular, the study of randomly perturbed labels dates back to \cite{angluin1988learning} in the context of random classification noise model. See, e.g., \cite{nettleton2010study, frenay2014classification} for a survey. Also, classification problems with missing labels are also extensively studied in the context of semi-supervised or weakly supervised learning by \cite{garcia2011degrees, joulin2012convex,patrini2016loss}, among others. Despite the extensive study on this problem, its information-theoretic and computational boundaries remain unexplored in terms of theory. In a nutshell, the information-theoretic boundary refers to the optimal statistical accuracy achievable by any algorithms, while the computational boundary refers to the optimal statistical accuracy achievable by the algorithms under a computational budget that has a polynomial dependence on the problem scale $(d, n)$. Moreover, it remains unclear how these two boundaries vary along~with $\alpha$. One interesting question to ask is how the degree of supervision affects the fundamental statistical and computational difficulties of this problem, especially in the high dimensional regime.


In this paper, we sharply characterize both the information-theoretic and computational boundaries of the weakly supervised binary classification problems under the minimax framework. Specifically, we consider the Gaussian generative model where $\bX | Z = z \sim \Gaussian(\bmu_z, \bSigma)$ and $z\in\{0,1\}$ is the true label. Suppose $\{(\xb_i, z_i)\}_{i=1}^n$ are $n$ independent samples of $(\bX, Z)$. We assume that $\{y_i\}_{i=1}^n$ are generated from $\{z_i\}_{i=1}^n$ in the aforementioned manner. We focus on the high dimensional regime, where $d \gg n$ and $\bmu_1 - \bmu_0$ is $s$-sparse, i.e., $\bmu_1 - \bmu_0 $ has $s$ nonzero entires. We are interested in estimating $\bmu_1 - \bmu_0$ from the observed samples $\{(\xb_i, y_i)\}_{i=1}^n$. By  a standard reduction argument \citep{tsybakov2008introduction}, the fundamental limits of this estimation task are captured by a hypothesis testing problem, namely, $H_0: \bmu_1 - \bmu_0 = \zero$ versus $H_1: \bmu_1 - \bmu_0$ is $s$-sparse and 
\#\label{eq:w1}
(\bmu_1 - \bmu_0)^\top \bSigma^{-1} (\bmu_1 - \bmu_0) := \gamma_n > 0,
\# 
where $\gamma_n$ denotes the signal strength that scales with $n$. Consequently, we focus on studying the fundamental limits of $\gamma_n$ for solving this hypothesis testing problem.

\begin{figure*}[!htb]
	\centering
	\includegraphics[width=0.95\textwidth]{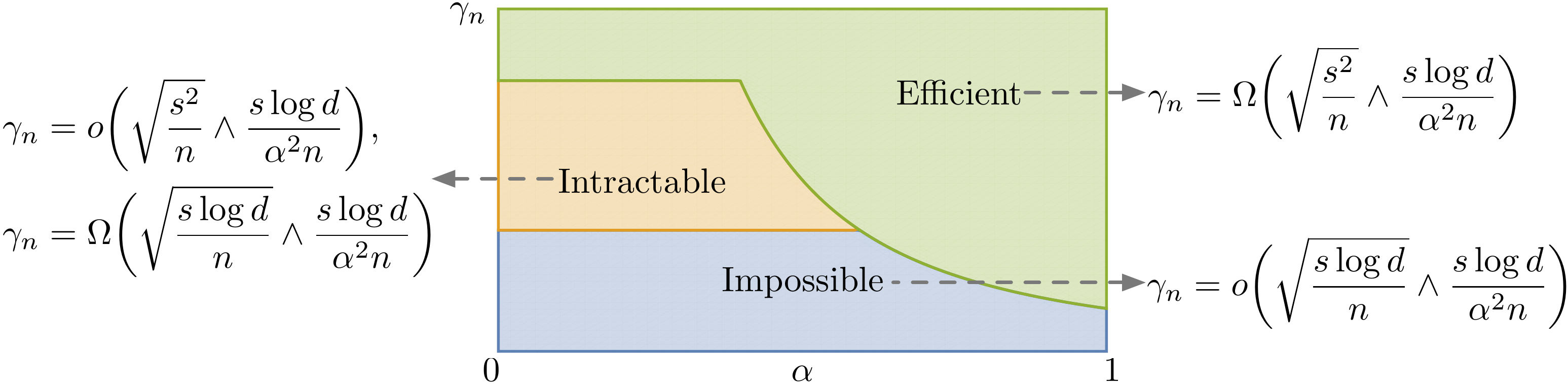}
	\caption{Computational-statistical phase transitions for weakly supervised binary classification.~Here $\alpha$ denotes the degree of supervision, i.e., the label is corrupted to be uniformly random with probability $1-\alpha$, and $\gamma_n$ is the signal strength, which is defined in \eqref{eq:w1}. Here $a \wedge b$ denotes $\min\{a, b\}$.}
	\label{fig:illu}
\end{figure*}
Our main results are illustrated in Figure \ref{fig:illu}. Specifically, we identify the impossible, intractable, and efficient regimes for the statistical-computational phase transitions under certain regularity conditions. 
\begin{enumerate}[label=(\roman*), wide, labelwidth=!, labelindent=0pt]
	\item For $\gamma_n = o[ \sqrt{{s \log d}/{n}} \wedge (1/\alpha^2\cdot{s\log d}/{n})]$, any algorithm is asymptotically powerless in solving the hypothesis testing problem. 
	\item For $\gamma_n = \Omega [  \sqrt{{s \log d}/{n}} \wedge (1/\alpha^2\cdot{s\log d}/{n})]$ and $\gamma_n = o[ \sqrt{{s^2}/{n}} \wedge (1/\alpha^2\cdot{s\log d}/{n})]$, any tractable algorithm that has a polynomial oracle complexity under an extension of the statistical query model \citep{kearns1998efficient} is asymptotically powerless.   We will rigorously define the computational model in \S\ref{sec:bg}. 
	\item For $\gamma_n = \Omega[ \sqrt{{s^2}/{n}} \wedge (1/\alpha^2\cdot{s\log d}/{n})]$, there is an efficient algorithm with a polynomial oracle complexity that is asymptotically powerful in solving the testing problem. 
\end{enumerate}
Here $\sqrt{{s \log d}/{n}} \wedge (1/\alpha^2\cdot{s\log d}/{n})$ gives the information-theoretic boundary, while $\sqrt{{s^2}/{n}} \wedge (1/\alpha^2\cdot{s\log d}/{n})$ gives the computational boundary. Moreover, by a reduction from the estimation problem to the testing problem, these boundaries for testing imply the ones for estimating $\bmu_2 - \bmu_1$ as well. 

Consequently, there exists a significant gap between the computational and information-theoretic boundaries for small $\alpha$. In other word, to achieve the information-theoretic boundary, one has to pay the price of intractable computation. As $\alpha$ tends to one, this gap between computational and information-theoretic boundaries narrows and eventually vanishes. This indicates that, having more supervision not only improves the statistical accuracy, as shown by the decay of information-theoretic boundary in Figure \ref{fig:illu}, but more importantly, enhances the computational efficiency by~reducing the computational price for attaining information-theoretic optimality. This phenomenon --- ``more supervision, less computation'' --- is observed for the first time in this paper. 

\subsection{More Related Work, Our Contribution, and Notation}
Besides the aforementioned literature on weakly supervised learning and label corruption, our work is also connected to a recent line of work on statistical-computational tradeoffs \citep{berthet2013computational, berthet2013optimal, chandrasekaran2013computational, ma2013computational, gao2014sparse, zhang2014lower, hajek2014computational, chen2014statistical, wang2014statistical,  wang2015sharp, fan2016curse}. In comparison, we quantify the  statistical-computational tradeoffs for weakly~supervised learning for the first time. Furthermore, our results are built on an oracle computational model in \cite{fan2016curse} that slightly extends the statistical query model \citep{kearns1998efficient}, and hence do not hinge on unproven conjectures on computational hardness like {\sf planted clique}. Compared with our work, \cite{fan2016curse} focuses on the computational hardness of learning   heterogeneous models, whereas we consider the   interplay between supervision and statistical-computational tradeoffs.   A similar computational model is used in  \cite{wang2015sharp} to study structural normal mean model and principal component analysis, which exhibit different statistical-computational phase transitions. In addition, our work is related to sparse linear discriminant analysis and two-sample testing of sparse means, which correspond to our special cases of $\alpha = 1$ and  $\alpha = 0$, respectively. See, e.g., \cite{fan2012road, tony2014two} for details. In contrast with their results, our results capture the effects of $\alpha$ on statistical and computational tradeoffs. 

In summary, the contribution of our work is two-fold: 
\begin{enumerate}[label=(\roman*), wide, labelwidth=!, labelindent=0pt]
	\item We characterize the computational and statistical boundaries of the weakly supervised binary classification problem for the first time. Compared with existing results for other models, our results do not rely on unproven conjectures. 
	\item Based on our theoretical characterization, we propose the ``more supervision, less computation'' phenomenon, which is observed for the first time. 
\end{enumerate}

\paragraph{Notation.} We denote the $\chi^2$-divergence   between two distributions $\mathbb{P},\mathbb{Q}$ by $\chisquare(\mathbb{P}, \mathbb{Q})$. For two nonnegative sequences $a_n, b_n$ indexed by $n$, we use $a_n = o(b_n)$ as a shorthand for $\lim_{n \rightarrow \infty} a_n/b_n = 0$. We say $a_n = \Omega(b_n)$ if  $a_n/b_n \geq c$ for some absolute constant $c > 0$ when $n$ is sufficiently large. We use $a \vee b$~and~$a \wedge b$ to denote~$\max\{a, b\}$~and~$\min\{a, b\}$,~respectively. For any positive integer $k$, we  denote $\{1, 2, \ldots, k\}$ by $[k]$. For~$\vb \in \RR^{d}$,~we denote by~$\| \vb\|_p$ the $\ell_p$-norm of $\vb$.~In addition, we denote the operator norm of a matrix~$\Ab$~by~$ \opnorm{\Ab}{2}$.

%% file: background_nips.tex

\section{Background}\label{sec:bg}
In this section, we formally define  the  statistical model for   weakly supervised binary classification. Then we follow it with the statistical  query model that connects computational complexity and  statistical optimality.

\subsection{Problem Setup}

Consider the following Gaussian generative model for binary classification. For a random vector $\bX \in \real^{d}$ and a binary random variable $Z \in \{0,1\}$, we assume
\begin{equation}\label{eq:model_1}
\bX|Z = 0 ~\sim~ \Gaussian(\meanv_0, \covMat),~~~~\bX|Z = 1 ~\sim~ \Gaussian(\meanv_1, \covMat),
\end{equation}
where $\prob(Z = 0) = \prob( Z = 1 ) = 1/2$. Under this model, the optimal classifier by Bayes rule corresponds to the Fisher's linear discriminative analysis (LDA) classifier. In this paper, we focus on the noisy label setting  where true label $Z$ is replaced by a uniformly random label in $\{0,1\}$ with probability $1 - \alpha$. Hence, $\alpha$ characterizes the degree of supervision in the model. In specific, if $\alpha = 0$,  we  observe  the true label $Z$, thus the problem belongs to supervised learning. Whereas if $\alpha = 1$, the observed label is completely random, which contains no information of the model in \eqref{eq:model_1}. This setting is thus equivalent to learning a Gaussian mixture model, which is an unsupervised problem.  In the general setting with noisy labels, we denote the observed label by $Y$, which is linked to the true label $Z$ via
\#\label{eq:model_2}
\prob( Y = Z) = (1+ \alpha) /2 ,~~ \prob( Y = 1 - Z) = (1 - \alpha) /2.
\#
We consider the hypothesis testing problem of detecting whether $\meanv_0 \ne \meanv_1$ given $n$ i.i.d. samples $\{y_i,\xb_i\}_{i=1}^n$ of $(Y,\bX)$, namely
\begin{equation} \label{eq:problem}
H_0\colon \meanv_0 = \meanv_1 ~~ \text{versus}~~ H_1\colon \meanv_0 \ne \meanv_1.
\end{equation}
We focus on the high dimensional and sparse regime, where $d \gg n$ and  $\meanv_0 - \meanv_1$ is $s$-sparse, i.e., $\meanv_0 - \meanv_1 \in \sparseSet(s)$, where $\sparseSet(s) := \{\meanv \in \real^{d} \colon  \|{\meanv}\|_0 \leq s\}$. Throughout this paper, use the sample size $n$ to drive the asymptotics. We introduce a shorthand notation $\btheta := (\meanv_0, \meanv_1, \covMat,  \alpha)$ to represent the parameters of the aforementioned model. Let $\prob_{\btheta}$ be the joint distribution of $(Y,\bX)$ under our statistical  model with parameter $\btheta$, and $\prob_{\btheta}^n$ be the product distribution of $n$ i.i.d. samples accordingly.  We denote  the parameter spaces  of the null and alternative hypotheses by  $\parSpace_0$ and $\parSpace_1$ respectively. For any test function $\phi: \{( y_i,\xb_i)\}_{i=1}^n \rightarrow \{0,1\}$, the classical testing risk is defined as the summation of type-I and type-II errors, namely
\$
\risk_n(\phi;\parSpace_0, \parSpace_1): = { \sup_{\btheta \in \parSpace_0}} \prob_{\btheta}^n ( \phi = 1  )  +{ \sup_{\btheta \in \parSpace_1}} \prob_{\btheta}^n ( \phi = 0).
\$
The  minimax risk is defined as the smallest testing risk of all possible  test functions, that is,
\#\label{eq::minimax_risk}
\risk^*_n(\parSpace_0, \parSpace_1) : = { \inf_{\phi}}~ \risk_n(\phi; \parSpace_0, \parSpace_1),
\#
where the infimum is taken over all measurable test functions.

Intuitively, the separation between two Gaussian components under $H_1$ and the covariance matrix $\bSigma$ together determine  the hardness of detection. To characterize such dependence, we define the signal-to-noise ratio (SNR) as $\snr(\btheta) := (\meanv_0  - \meanv_1)^{\top}\covMat^{-1}(\meanv_0- \meanv_1)$. For any nonnegative sequence~$\{ \gamma_n \}_{n \geq 1}$, let $\parSpace_1(\gamma_n) := \{\btheta : \snr(\btheta) \geq \gamma_n\}$~be a sequence of alternative parameter spaces   with minimum separation~$\gamma_n$. The following minimax rate  characterizes the information-theoretic  limits of the detection problem.
\begin{definition} [Minimax rate] \label{def:minimax_rate} We say a sequence $\{ \gamma_n^*\}_{n \geq 1}$ is a minimax rate if 
	\begin{itemize}[ wide, labelwidth=!, labelindent=0pt]
		\item For any sequence $\{ \gamma_n\}_{n \geq 1}$ satisfying $\gamma_n = o(\gamma_n^*)$, we have $\lim_{n \rightarrow \infty} \risk^*_n [ \parSpace_0, \parSpace_1(\gamma_n)] =  1$;
		\item For any sequence $\{ \gamma_n\}_{n \geq 1}$ satisfying $\gamma_n = \Omega(\gamma_n^*)$, we have $\lim_{n \rightarrow \infty} \risk^*_n [\parSpace_0, \parSpace_1(\gamma_n)]  = 0$.
	\end{itemize}
\end{definition}

The minimax rate in Definition \ref{def:minimax_rate} characterizes the statistical difficulty of the testing problem. However, it fails to shed light on the computational efficiency of possible testing algorithms. The reason is that this concept does not make any computational restriction on the test functions. The minimax risk in \eqref{eq::minimax_risk} might be attained only by test functions that have exponential computational complexities. This limitation of Definition \ref{def:minimax_rate} motivates us to study statistical limits under computational constraints.


\subsection{Computational Model}\label{sec::oracle_model}
Statistical query models \citep{kearns1998efficient, feldman2013statistical, feldman2015statistical, feldman2015complexity, wang2015sharp, fan2016curse} capture computational complexity by characterizing the total number of rounds an algorithm interacts with data. In this paper, we consider the following statistical query model, which admits bounded query functions but allows the responses of query functions to be unbounded.
\begin{definition}[Statistical query model]\label{def::oracle}
	In the statistical query model,  an 
	algorithm $\mA$ is allowed to query an oracle $T$ rounds, but not to access data~$\{( y_i, \xb_i)\}_{i=1}^n$ directly.  At each round, $\mA$ queries the oracle $r$ with a query function~$q \in \cQ_{\mA}$, in which $\cQ_{\mA} \subseteq \{q: \{ 0,1\}\times \RR^d \rightarrow [ -M, M]\}$ denotes the query space of $\mA$. The oracle $r$ outputs  a realization of a random variable $Z_q \in \RR$ satisfying 
	\# 
	& \PP \biggl( \bigcap_{q \in \cQ_{\mA}} \bigl \{    | Z_q  - \EE[q(Y, \bX)]    |  \leq  \tau_q  \bigr\} \biggr) \geq 1- 2\xi,~~\text{where~~}  \notag\\
	&\tau_q =    [\eta(\cQ_{\mA}) + \log (1/\xi)   ]  \cdot M/n \bigvee \sqrt{2   [\eta(\cQ_{\mA}) + \log (1/\xi) ] \cdot  (M^2 -  \{  \EE [ q( Y, \bX) ]  \}^2    )  \big/n}. \label{eq::query_2}
	\#
	Here $\tau_q >0$ is the  tolerance parameter and $\xi\in [0,1)$ is the tail probability.  The quantity $\eta(\cQ_{\mA}) \geq 0$  in $\tau_q$ measures~the capacity of $\cQ_{\mA}$ in logarithmic scale, e.g., for countable $\cQ_{\mA}$, $\eta(\cQ_{\mA}) = \log (|\cQ_{\mA}|)$. The number $T$ is defined as the oracle complexity. We denote by   $\cR[\xi,n,T, \eta(\cQ_{\mA}) ]$  the set of oracles satisfying \eqref{eq::query_2}, and by $\cA(T)$ the family of algorithms that queries an oracle no more than $T$~rounds.
\end{definition}

This version of statistical query model is used in \cite{fan2016curse}, and reduces to the VSTAT model proposed in \cite{feldman2013statistical, feldman2015statistical, feldman2015complexity} by the transformation $\tilde q(y, \xb) = q(y, \xb) / (2M) + 1/2$ for any $q \in \cQ_{\mA} $. The computational model in Definition \ref{def::oracle} enables us to handle query functions that are bounded by an unknown and fixed number $M$. Note that that by incorporating the tail probability $\xi$, the response $Z_q$ is allowed to be unbounded.  To understand the  intuition behind  Definition \ref{def::oracle},~we remark  that   \eqref{eq::query_2} resembles the Bernstein's inequality for bounded  random variables~\citep{vershynin2010introduction}
\#\label{eq::bernstein}
\PP\biggl \{  \biggl | \frac{1}{n} \sum_{i=1}^n   q(Y_i, \bX_i)  - \EE[q(Y, \bX)] \biggr | \geq    t  \biggr\} \leq 2 \exp   \biggl \{  \frac{t^2 } {2 \text{Var} [q(Y, \bX)  ] + M t}   \biggr\}.
\#
We first replace $\text{Var}\  [q(Y, \bX)   ] $ by its upper bound  $M^2 -   \{  \EE [ q( Y, \bX) ]   \}^2 $, which is tight when $q$ takes values in $\{ -M, M \}$. Then inequality~\eqref{eq::query_2}~is obtained by replacing~$ n^{-1}    \sum_{i=1}^n  q(Y_i, \bX_i)$ in~\eqref{eq::bernstein} by $Z_{q}$ and then bounding the suprema over the query space $\cQ_{\mA}$. In the definition of 
 $\tau_q$ in \eqref{eq::query_2}, we incorporate the effect of uniform concentration over the query space $\cQ_{\mA}$ by adding the quantity $\eta(\cQ_{\mA})$, which measures the capacity of $\cQ_{\mA}$.  In addition, under the Definition \ref{def::oracle}, the algorithm $\mA$   does not interact directly with data.  Such an restriction characterizes the fact that in statistical problems, the  effectiveness of an algorithm only depends on the global statistical   properties,  not the information of individual data points. For instance, algorithms that only rely on the convergence of the empirical distribution to the population distribution are contained in the statistical query model; whereas algorithms that hinge on the first data point $(y_1, \xb_1)$ is not allowed.  This restriction captures a vast family of  algorithms in statistics and machine learning, including applying gradient method to maximize likelihood function, matrix factorization algorithms, expectation-maximization algorithms, and sampling algorithms \citep{feldman2013statistical}.

Based on the statistical query  model,   we study the minimax risk under oracle complexity constraints.  For the testing problem \eqref{eq:problem},  let~$\mathcal{A}(T_n)$~be a class of testing algorithms  under the statistical query model with query complexity   no more than~$T_n$, with $\{ T_n\}_{n\geq 1}$ being a sequence of positive integers depending on the sample size $n$. For any $\mA \in \cA(T_n)$ and any oracle~$r \in\cR[\xi,n,T_n, \eta(\cQ_{\mA}) ]$ that responds to $\mA$, let $\cH(\mA,r)$ be the set of  test functions that deterministically depend on $\mA$'s queries to the oracle $r $ and~the corresponding  responses.  We use $\overline{\PP}_{\btheta}$ to denote the distribution of the random variables returned by   oracle $r$ when the model  parameter is $\btheta$.

For a general hypothesis testing problem, namely, $H_0\colon \btheta \in \cG_0$ versus $H_1 \colon \btheta\in \cG_1$, the minimax testing~risk~with respect to an algorithm $\mA$ and a statistical oracle $r \in\cR[\xi,n,T_n, \eta(\cQ_{\mA}) ]$  is defined~as
\begin{equation}\label{eq::minimax_risk_oracle}
\overline{R}_n^*(\cG_0, \cG_1;\mA, r) : = { \inf_{\phi\in \cH(\mA,r)}}  \biggl  [ { \sup _{\btheta \in \cG_0}} \overline{\PP}_{\btheta}(\phi = 1) + { \sup _{\btheta \in \cG_1}} \overline{\PP}_{\btheta}(\phi = 0) \biggr ].
\end{equation}
Compared with the classical   minimax   risk in \eqref{eq::minimax_risk}, the new notion in   \eqref{eq::minimax_risk_oracle} incorporates the~computational budgets via oracle complexity. In specific, we only consider the test functions obtained by an algorithm  with  at most  $T_n$ queries to a statistical oracle. If $T_n$ is a polynomial of  the dimensionality  $d$, \eqref{eq::minimax_risk_oracle} characterizes the statistical optimality of computational efficient algorithms.
This motivates us to define the computationally tractable minimax rate, which contrasts with Definition~\ref{def:minimax_rate}. 
\begin{definition} [Computationally tractable minimax rate]  \label{def::comp_minimax_rate}  Let $\parSpace_1(\gamma_n) := \{\btheta : \snr(\btheta) \geq \gamma_n\}$ be a sequence of model spaces with minimum separation $\gamma_n$, where $\rho(\btheta)$ is  the SNR. A sequence $\{ \overline{\gamma}_n^*\}_{ n \geq 1}$ is called a computationally tractable minimax rate  if 
	\begin{itemize}[wide, labelwidth=!, labelindent=0pt]
		\item For any sequence $\{ \gamma_n  \}_{n \geq 1} $ satisfying $\gamma_n = o(\overline{\gamma}_n^*)$,  any constant $\eta > 0$,  and any $\mA \in \cA(d^\eta )$,  there exists an oracle $r \in \cR[\xi,n,T_n, \eta(\cQ_{\mA}) ]$ such that   $\lim_{n \rightarrow \infty} \overline{\risk}^*_n [ \parSpace_0, \parSpace_1(\gamma_n); \mA, r ] =  1$;
		\item For any sequence $\{ \gamma_n  \}_{n \geq 1} $~satisfying $\gamma_n = \Omega(\overline{\gamma}_n^*)$, there exist  a constant $\eta >0$ and an algorithm $\mA \in \cA(d^{\eta})$ such that, for any $r \in  \cR[\xi,n,T_n, \eta(\cQ_{\mA}) ]$, we have $\lim_{n \rightarrow \infty} \overline{\risk}^*_n [ \parSpace_0, \parSpace_1(\gamma_n); \mA, r ] = 0$.
	\end{itemize}
\end{definition}

%% file: main_results_nips.tex

\section{Main Results}\label{sec::main_results}
Throughout this paper, we assume that the covariance matrix  $\bSigma$  in \eqref{eq:model_1} is known. Specifically, for some positive definite $\bSigma\in \RR^{d \times d}$, the parameter spaces of the null and alternative hypotheses are defined as
\# 
\parSpace_0(\bSigma) &:= \{\btheta = (\meanv, \meanv, \bSigma, \alpha): \meanv \in \real^d\},\label{eq:G_0} \\
\parSpace_1(\bSigma; \gamma_n)& := \{\btheta = (\meanv_0, \meanv_1, \bSigma, \alpha): \meanv_0,\meanv_1 \in \real^d, \meanv_0 - \meanv_1 \in \sparseSet(s), \snr(\btheta) \geq \gamma_n\}. \label{eq:G_1}
\#
Accordingly, the testing problem of detecting whether $\bmu_0 \neq \bmu_1$ is to distinguish 
\#\label{eq:problem2}
H_0 \colon \btheta \in \cG_0(\bSigma)~~\text{versus}~~H_1 \colon \btheta \in\cG_1(\bSigma; \gamma_n).
\#
In \S \ref{sec:info_limits}, we present the minimax rate of the detection problem from an information-theoretic  perspective. In \S\ref{sec::comp_limits}, under the statistical query model introduced in \S \ref{sec::oracle_model}, we provide a computational lower bound  and  a nearly matching upper bound that is achieved by an efficient testing algorithm.

\subsection{Information-theoretic Limits} \label{sec:info_limits}
Now we turn to characterize the minimax rate given in Definition \ref{def:minimax_rate}. For parameter spaces \eqref{eq:G_0} and \eqref{eq:G_1} with known $\bSigma$, we show that in highly sparse setting where $s = o(\sqrt{d})$, we have
\#\label{eq::main_result1}
\gamma_n^* = \sqrt{ s\log d / n} \wedge   (  1/ \alpha^2  \cdot s\log d/   n),
\#
To prove \eqref{eq::main_result1}, we first present a lower bound which shows that the hypothesis testing  problem in \eqref{eq:problem2} is impossible if $\gamma_n  =o(\gamma_n^*)$.

\begin{theorem} \label{thm:info_lower_bound} For the  hypothesis testing  problem in \eqref{eq:problem2} with known $\bSigma$,  we assume that there exists a small constant $\delta >0$ such that   $s = o(d^{1/2 - \delta})$. Let $\gamma_n^*$ be defined in \eqref{eq::main_result1}.  For any sequence $\{ \gamma_n\}_{ n \geq 1}$ such that  $\gamma_n = o(\gamma_n^*)$,  any hypothesis test is asymptotically powerless, namely,
	$$
	\lim_{n \rightarrow \infty}~ \sup_{\bSigma} \risk_n^*[ \parSpace_0(\bSigma), \parSpace_1(\bSigma ;\gamma_n)]  =  1.
	$$	
\end{theorem}

By Theorem \ref{thm:info_lower_bound}, we observe a phase transition  in the necessary SNR for powerful detection  when $\alpha $ decreases from one  to zero. Starting with rate~$s\log d / n$ in  the supervised setting where~$\alpha = 1$, the required  SNR gradually increases as label qualities decrease. Finally, when $\alpha$ reaches zero, which corresponds to the unsupervised setting, powerful detection requires the SNR to be $\Omega(\sqrt{s \log d/n})$. 
It is worth noting that  when $\alpha =  (s\log d/n)^{1/4} $, we still have $( n ^3 s \log d)^{1/4}$~uncorrupted labels. However, our lower bound (along with the upper bound shown in Theorem \ref{thm:info_upper_bound}) indicates that the information contained in these uncorrupted labels are buried in the noise, and cannot essentially improve the detection quality compared with the unsupervised setting.

Next we establish a  matching upper bound for the detection problem in \eqref{eq:problem2}.
We denote the condition number of  the covariance matrix  $\bSigma$ by $\kappa$, i.e., $\kappa := \lambda_{\text{max}}(\bSigma)/\lambda_{\text{min}}(\bSigma)$,
where $\lambda_{\text{max}}(\bSigma)$ and $\lambda_{\text{min}}(\bSigma)$ are the largest and smallest eigenvalues of $\bSigma$, repectively. Note that marginally $Y$ is uniformly distributed over $\{0,1\}$.  
For ease of presentation, we assume that the sample size is $2n$ and each class contains exactly $n$ data points. Note that we can always discard some samples in the larger class to make the sample sizes of both classes to be equal. Due to the law of large numbers, this trick will not affect the analysis of sample complexity in the sense of order wise.

Given $2n$ i.i.d. samples $\{( y_i, \xb_i)\}_{i=1}^{2n}$ of $(Y, \bX)\in \{0,1\} \times  \real^d $, we define  
\#\label{eq::compute_diff_sample_1}
\wb_i = \bSigma^{-1/2}(\xb_{2i} - \xb_{2i - 1}),~ \text{for all} ~i \in [n].
\#
In addition,   we split the dataset $\{( y_i, \xb_i)\}_{i=1}^{2n}$ into two disjoint parts $\{( 0, \xb^{(0)}_{i})\}_{i=1}^{n }$ and $\{(1, \xb^{(1)}_{i})\}_{i=1}^{n }$, and define  
\#\label{eq::compute_diff_sample_2}
\ub_i =  \xb^{(1)}_{i} - \xb^{(0)}_{i}, ~ \text{for all} ~i \in [n].
\#
We note that computing sample differences in \eqref{eq::compute_diff_sample_1} and  \eqref{eq::compute_diff_sample_2} is critical for our problem because we focus on detecting the difference between $\meanv_0$ and $\meanv_1$, and computing differences can avoid estimating  $\EE_{\PP_{\btheta}} (\bX)$  that might be dense. For any integer $s \in [d]$, we define  $\mathcal{B}_2(s) := \sparseSet(s) \cap \mathbb{S}^{d-1}$ as the set of $s$-sparse vectors on the unit sphere in $\RR^d$. With $\{\wb_i\}_{i=1}^{n}$ and $\{\ub_i\}_{i=1}^{n}$, we introduce two test functions 
\#
\phi_1 := \ind\biggl \{\sup_{\vb \in \mathcal{B}_2(s)}  \frac{1}{n}\sum_{i=1}^{ n }\frac{(\vb^{\top}\bSigma^{-1}\wb_i)^2}{2\vb^{\top}\bSigma^{-1}\vb} \geq 1 + \tau_1  \biggr\},\label{eq:test_1} \\
\phi_2 := \ind\biggl \{\sup_{\vb \in \mathcal{B}_2(1)}  \frac{1}{n}\sum_{i=1}^{n} \langle{\vb}, {\diag(\bSigma)^{-1/2}\ub_i}\rangle \geq \tau_2  \biggr\},\label{eq:test_2}
\#
where $\tau_1, \tau_2 > 0$ are algorithmic parameters that will be specified later.  To provide some intuitions, we consider the case where $\bSigma = \Ib$. Test function $\phi_1$ seeks a sparse direction that explains the most variance of $\wb_i$. Therefore, such a test is closely related to the sparse principal component detection problem \citep{berthet2013optimal}. Test function $\phi_2$ simply selects the coordinate of $n^{-1}    \sum_{i=1}^n \ub_i$ that has the largest magnitude and compares it with $\tau_2$. This test is closely related to detecting sparse normal mean in high dimensions \citep{johnstone1994minimax}. Based on these two ingredients, we construct our final testing function $\phi$ as $\phi = \phi_1 \vee \phi_2$, i.e., if any of $\phi_1$ and $\phi_2$ is true, then $\phi$ rejects the null. The following theorem  establishes a sufficient condition for  test function $\phi$ to be asymptotically powerful.
\begin{theorem} \label{thm:info_upper_bound}
	Consider the testing problem \eqref{eq:problem2} where $\bSigma$ is known and    has condition number $\kappa$. 
	For  test functions $\phi_1$ and $\phi_2$  defined in   \eqref{eq:test_1} and \eqref{eq:test_2} with   parameters $\tau_1$ and $\tau_2$  given by
	$$
	\tau_1 = \kappa\sqrt{ s\log(ed/s)/ n },~~ \tau_2 =  \sqrt{ 8 \log d/ n}.
	$$
	We define the ultimate test function as  $\phi = \phi_1 \vee \phi_2$.  We assume that $ s \leq C\cdot  d $ for some absolute constant $C s$  and 
	$n \geq 64\cdot s\log(ed/s)$.  Then if 
	\#\label{eq:snr_upper_bound}
	\gamma_n \geq C'\kappa\cdot  [ \sqrt{ s\log(ed/s)/ n} \wedge  ( 1/ \alpha^2\cdot s \log d/ n  )  ],
	\#
	where $C'$ is an absolute constant, then test function $\phi$ is asymptotically powerful. In specific, we have
	\#\label{eq:risk_upper_bound}
	\sup_{\btheta \in \parSpace_0(\bSigma)} \prob_{\btheta}^n( \phi = 1 )  +  \sup_{\btheta \in \parSpace_1(\bSigma; \gamma_n)} \prob_{\btheta}^n ( \phi = 0) \leq 20/d.
	\#
\end{theorem}

Theorem \ref{thm:info_upper_bound} provides a non-asymptotic guarantee. When $n$ goes to infinity, \eqref{eq:risk_upper_bound} implies that  the test function $\phi$ is asymptotically powerful. When $s = o(\sqrt{d})$ and $\kappa$ is a constant, \eqref{eq:snr_upper_bound} yields  $\gamma_n = \Omega[\sqrt{s\log d/n}\wedge ( 1/ \alpha^2 \cdot  s\log d/  n)]$, which matches the lower bound given in Theorem \ref{thm:info_lower_bound}. Thus we conclude that $\gamma_n^*$ defined in \eqref{eq::main_result1}  is the minimax rate of testing problem in \eqref{eq:problem2}.  We remark that when $s = \Omega(d)$, $\alpha = 1$, i.e., the standard (low-dimensional) setting of two sample testing, the bound provided in \eqref{eq:snr_upper_bound} is sub-optimal as \cite{ramdas2016classification} shows that SNR rate $\sqrt{d}/n$ is sufficient for asymptotically powerful detection when $n = \Omega(\sqrt{d})$. It is thus worth noting that we focus on the highly sparse setting $s = o(\sqrt{d})$ and provided sharp minimax rate for this regime. In the definition of $\phi_1$ in \eqref{eq:test_1}, we search over the set $\cB_2(s)$. Since $\cB_2(s)$ contains ${d\choose s}$ distinct sets of supports, computing $\phi_1$ requires exponential running time. 


\subsection{Computational Limits}\label{sec::comp_limits}
In this section, we characterize the computationally tractable minimax rate $\overline{\gamma}_n^*$ given in Definition~\ref{def::comp_minimax_rate}. Moreover, we focus on the setting where $\bSigma$ is known a priori and the parameter spaces for the null and alternative hypotheses are defined in   \eqref{eq:G_0} and~\eqref{eq:G_1}, respectively.  The main result is that, in highly sparse setting where $s = o(\sqrt{d})$,  we have
\begin{equation}\label{eq::main_result2}
\overline{\gamma}_n^* = \sqrt{s^2 /  n } \wedge (1 /\alpha^2 \cdot   s \log d / n).
\end{equation}
We first present the lower bound in the next result.

\begin{theorem}\label{thm::comp_limit}
	For the   testing  problem in \eqref{eq:problem2} with $\bSigma$  known a priori,  we make the same assumptions as in Theorem \ref{thm:info_lower_bound}.  For any sequence $\{ \gamma_n\}_{ n \geq 1}$ such that  \#\label{eq:real_comp}
	\gamma_n = o  \left\{  \gamma_n ^*\vee   \left[ \sqrt{s^2 /  n } \wedge (1 /\alpha^2 \cdot   s  / n)   \right]  \right \} , \#  where  $\gamma_n^*$ is defined in \eqref{eq::main_result1}, any computationally tractable test is asymptotically powerless under the statistical query model. That is, 
	for any constant $\eta > 0$ and any $\mA \in \cA( d^{\eta} )$, there exists an oracle $r \in \cR[\xi,n,T_n, \eta(\cQ_{\mA}) ] $  such that  $\lim_{n \rightarrow \infty} \overline{\risk}^*_n [ \parSpace_0( \bSigma) , \parSpace_1(\bSigma, \gamma_n); \mA, r ] =   1.$	
\end{theorem}

We remark that the lower bound in \eqref{eq:real_comp} differs from $\gamma_n^*$ in \eqref{eq::main_result2} by a logarithmic term when $\sqrt{1/n} \leq \alpha^2 \leq \sqrt{s \log d/n}$.  We expect this gap to be eliminated by more delicate analysis under the statistical query model.

Now putting   Theorems \ref{thm:info_lower_bound} and \ref{thm::comp_limit} together, we describe the ``more supervision, less computation'' phenomenon as follows.
\begin{enumerate}[label=(\roman*), wide, labelwidth=!, labelindent=0pt]
	\item When $0\leq \alpha \leq (\log ^2 d /n)^{1/4}$,   the computational lower bound implies that the  uncorrupted labels are unable to improve the quality of computationally tractable detection compared with the unsupervised setting. In addition, in this region, the gap between   $\gamma_n^*$ and $\overline \gamma_n^*$ remains the same.
	\item When $ (\log ^2 d /n)^{1/4}< \alpha \leq (s \log d /n)^{1/4}$, the information-theoretic lower bound shows that the uncorrupted labels cannot improve the quality of detection compared with unsupervised setting. However,   more uncorrupted labels improve  the statistical performances of hypothesis tests that are computationally tractable by shrinking the gap between $\gamma_n^* $ and $\overline \gamma_n^*$. 
	\item When $(s \log d /n)^{1/4} < \alpha \leq 1$,  having more uncorrupted labels improves both statistical optimality and the computational efficiency. In specific, in this case, the gap between $\gamma_n^* $ and $\overline \gamma_n^*$ vanishes and we have $\gamma_n^* = \overline \gamma_n^* = 1/ \alpha^2 \cdot s\log d/n$.
\end{enumerate}

Now we   derive a nearly matching upper bound under the statistical query model, which establishes the computationally tractable minimax rate together with Theorem \ref{thm::comp_limit}.
We construct a computationally efficient  testing procedure that combines two test functions which yields the two parts in $\overline \gamma_n^*$ respectively.  Similar to  $\phi_1$ defined in \eqref{eq:test_1}, the first test function discards the information of labels, which works for the purely unsupervised setting where $\alpha = 0$. For $j \in [d]$, we denote by $\sigma_{j}$ the $j$-th diagonal element of $\bSigma$.  Under the statistical query model, we consider the  $2d$ query functions
\#
&q_j (y, \xb) := x_j / \sqrt{\sigma_j} \cdot \ind\{ |x_j  /\sqrt{\sigma_j}| \leq R \cdot \sqrt{\log d}  \}  , \label{eq::query1} \\
& \tilde {q}_j(y, \xb) := (x_j^2 /\sigma_j - 1)\cdot \ind\{ |x_j /\sqrt{\sigma_j} | \leq R \cdot \sqrt{\log d}\} , ~\text{for~all~} j \in [d],\label{eq::query12}
\#

where $ R >0$ is an absolute constant.  Here we apply truncation to the query functions to obtain bounded queries, which is specified by the  statistical query model in Definition \ref{def::oracle}.
We denote by $z_{q_j}$ and $z_{\tilde{q}_j}$ the realizations of the random variables output by the statistical oracle for query functions $q_j$ and $\tilde q_j$, respectively. As for the second test function, similar to \eqref{eq:test_2}, we  consider 
\#\label{eq::query2}
\overline{q}_\vb (y, \xb) =   (2 y -1) \cdot \vb^\top  \text{diag}(\bSigma)^{-1/2} \xb \cdot \ind \bigl \{ | \vb^\top  \text{diag}(\bSigma)^{-1/2} \xb | \leq R \cdot \sqrt{\log d}\bigr \} 
\#
for all $\vb \in \mathcal{B}_2(1)$.
We denote by $Z_{\overline{q}_{\vb}} $ the output of  the statistical oracle corresponding to query function $\overline{q}_{\vb}$.
With these $4d$ query functions,  we introduce test functions
\#
\overline {\phi}_1&:= \ind \biggl \{\sup_{ j  \in [d]} (z_{\tilde{q}_j} - z_{q_j}^2) \geq    C  \overbar{\tau} _1 \biggr\}, ~~
\overline {\phi}_2  := \ind\biggl \{\sup_{\vb \in \cB_2(1)}    z_{\overline{q}_{\vb} } \geq  2\overline{\tau} _2\biggr\},\label{eq::test4}
\#
where $\overline{\tau}_1 $  and $\tau_2$ are    positive parameters that will be  specified later and $C$ is an absolute constant.
 
\begin{theorem}\label{thm::comp_upper}
	For the test functions $\overline \phi_1$ and $\overline \phi_2$ defined in \eqref{eq::test4} , we define the ultimate test function  as $\overline{\phi} = \overline{\phi}_1 \vee \overline {\phi}_2$.  We set
	\#\label{eq:trunc_tol}
	\overline  \tau_1  =R^2  \log d \cdot    \sqrt{\log (4 d / \xi)/n} , ~~ \overline\tau_2 = R \sqrt{\log d} \cdot    \sqrt{\log (4 d / \xi)/n},
	\#
	 where  $\xi = o(1)$. For the hypothesis testing problem in \eqref{eq:problem2},  we further assume that   $\| \bmu_0\| _{\infty} \vee \| \bmu_1 \|_{\infty} \leq C_0$ for some constant $C_0>0$.~Under the assumption that 
	\#\label{eq::computational_upper_bound}
	  \sup_{j\in [d]}\, (\mu_{0,j} - \mu_{1, j} )^2 / \sigma_{j}  = \Omega  \left\{    \left [   1/ \alpha^2 \cdot \log ^2 d \cdot \log ( d / \xi)/ n   \right]  \wedge  \log d\cdot \sqrt{\log( d/\xi)/ n} \right\}, 
	\#
	the risk of $\overline \phi$ satisfies that 
	$
	\overline{R}_n ^*  (\overline{\phi}) = \sup_{\btheta \in \cG_0(\bSigma) } \overline{\PP}_{\btheta} ( \overline{\phi} = 1) + \sup_{\btheta \in \cG_1(\bSigma, \gamma_n)} \overline{\PP}_{\btheta} ( \overline{\phi} = 0)\leq 5 \xi.
	$
	Here we denote by  $\mu_{0,j }$ and $\mu_{1, j}$ the $j$-th entry of $\bmu_0$ and $\bmu_1$, respectively.
\end{theorem}
\vskip-5pt
If we set the tail probability of the statistical query model to be  $\xi = 1/ d$, \eqref{eq::computational_upper_bound} shows  that  $\overline \phi$ is asymptotically powerful if ${ \sup_{j\in [d]}} ( \mu_{0,j} - \mu_{1, j} )^2 / \sigma_{j}  = \Omega [ (  1/ \alpha^2 \cdot \log^3  d / n )  \wedge  (\log^3  d / n)^{1/2} ] $. When the energy of $\bmu_0 - \bmu_1$ is spread over its support, $\| \bmu_0 - \bmu_1\|_{\infty}$ and $\| \bmu_0 - \bmu_1 \|_2 / \sqrt{s}$ are close.  Under the assumption that the condition number $\kappa$ of $\bSigma$ is a constant, \eqref{eq::computational_upper_bound} is implied by 
$$
{\gamma}_n \gtrsim (s^2 \log^3 d/  n )^{1/2} \wedge (1 /\alpha^2 \cdot   s\log^3 d/ n).
$$
Compared with Theorem  \ref{thm::comp_limit}, the above upper bound matches the  computational lower bound  up to a logarithmic factor and $\overline \gamma_n^*$ is between $\sqrt{s^2  /  n } \wedge (1 /\alpha^2 \cdot   s\log d/ n)$ and $( s^2 \log ^3 d/  n )^{1/2}  \wedge (1 /\alpha^2 \cdot   s\log^3  d/ n)$.  Note that  the  truncation on query functions in \eqref{eq::query1} and \eqref{eq::query12} yields an additional logarithmic term, which could be reduced to $( s^2 \log  d/  n )^{1/2}  \wedge (1 /\alpha^2 \cdot   s\log  d/ n)$ using more delicate analysis.
Moreover, the test function $\overline \phi_1$ is essentially  based on a diagonal thresholding algorithm performed on the covariance matrix of $\bX$.  The work in \cite{deshpande2014sparse} provides a more delicate analysis of this algorithm which establishes the $\sqrt{s ^2 /n}$ rate. Their algorithm can also be formulated into the statistical query model; we use the simpler version in \eqref{eq::test4} for ease of presentation. Therefore, with more sophicated proof techinique, it can be shown that $\sqrt{s^2  /  n } \wedge (1 /\alpha^2 \cdot   s\log d/ n)$ is the critical threshold for asymptotically powerful detection with computational efficiency. 
\subsection{Implication for Estimation}
Our aforementioned phase transition in the detection problems directly implies the statistical and computational trade-offs in the problem of estimation. We consider the problem of estimating the parameter  $\Delta \bmu = \bmu_0 - \bmu_1$ of   the binary classification model in  \eqref{eq:model_1} and \eqref{eq:model_2}, where $\Delta \bmu $ is $s$-sparse and  $\bSigma$ is  known a priori.  We assume that the signal to noise ratio is $\rho(\btheta) = \Delta \bmu^\top \bSigma ^{-1} \Delta \bmu \geq \gamma_n = o(\overline \gamma_n^*)$.  For any constant $\eta >0$ and any $\mA \in \cA(T)$ with $T = O(d^\eta)$, suppose we obtain an estimator $ \Delta  \hat \bmu $ of $\Delta \bmu$ by algorithm~$\mA$ under the statistical query model.  If $\Delta \hat \bmu$ converges to $\Delta \bmu$ in the sense~that 
\$
(\Delta \hat \bmu    - \Delta \bmu)^{\top} \bSigma^{-1} ( \Delta\hat  \bmu - \Delta \bmu)   = o[ \gamma_n^2 / \rho(\btheta) ],
\$
we have $| \Delta \hat \bmu ^\top \bSigma^{-1} \Delta \hat \bmu - \Delta \bmu^\top \bSigma ^{-1} \Delta \bmu | = o( \gamma_n)$. Thus the test function $\phi = \ind\{\Delta \hat \bmu ^\top \bSigma \Delta \hat \bmu \geq \gamma_n /2\}$  is asymptotically powerful, which contradicts the computational lower bound  in Theorem \ref{thm::comp_limit}. Therefore, there exists a constant $C$ such that $(\Delta \hat \bmu    - \Delta \bmu)^{\top} \bSigma^{-1} ( \Delta\hat  \bmu - \Delta \bmu)  \geq C \gamma_n^2 / \rho(\btheta)$ for any estimator $\Delta \hat \bmu$ constructed from polynomial number of queries.


%
%

%% file: proof.tex

\appendix{}
\section{Proofs of the Main Results}

\subsection{Proof of Theorem \ref{thm:info_lower_bound}} \label{proof:thm:info_lower_bound}
In this section, we  prove the information-theoretic  lower bound.  In specific, we focus on the restricted testing problem
\#\label{eq::restricted_test}
H_0 : \btheta = (\bm{0}, \bm{0}, \Ib, \alpha) ~~\text{versus}.~~ H_1 : \btheta = (-\vb/2, \vb/2, \Ib, \alpha),
\#
where 
\$
\vb \in \mathcal{H}(s) := \{\ub \in \{0,\beta\}^d \colon \zeronorm{\ub} = s\}.
\$
 Here we set $s \beta^2 = \gamma_n$ to ensure that $(-\vb/2, \vb/2, \Ib, \alpha)$  belongs to the alternative parameter  space $\cG (\bSigma; \gamma_n)$. For notational simplicity, we  denote the distribution of model $(-\vb/2, \vb/2, \Ib, \alpha)$ by $\prob_{\vb}$ and the product distribution of $n$ i.i.d. samples by $\prob_{\vb}^n$. By the definition of the minimax risk in \eqref{eq::minimax_risk},  we have 
\$
 \sup_{\bSigma} \risk_n^*\left [ \parSpace_0(\bSigma), \parSpace_1(\bSigma ;\gamma_n) \right] \geq \inf_{\phi}\left[\prob_{\bm{0}}^n(\phi = 1) +\frac{1}{ |\mathcal{H}(s)|}   \sum_{\vb \in \mathcal{H}(s) } \prob_{\vb}^n( \phi = 0) \right].
\$
We thus  reduce the minimax risk to the risk of a simple-against-simple hypothesis test where the alternative hypothesis corresponds to a uniform mixture of $\{ \PP_{\vb} \colon \vb \in \cH(s)\}$.  For notational simplicity, we define $\prob_{\mathcal{H}}^n := 1/ |\mathcal{H}(s)| \cdot  \sum_{\vb \in \mathcal{H}(s) }\prob_{\vb}^n$.   By Neyman-Pearson Lemma, we have
\[
\risk^*_n\left[ \parSpace_0, \parSpace_1(\bSigma; \gamma_n)\right] \geq 1 - \TV(\prob_{\bm{0}}^n, \prob_{\mathcal{H}}^n).
\]
Using  Pinsker's inequality  $\TV(\prob_{\bm{0}}^n, \prob_{\mathcal{H}}^n) \leq \sqrt{\chisquare(\prob_{\mathcal{H}}^n, \prob_{\bm{0}}^n)}$, for showing $\risk^*_n[\parSpace_0(\bSigma) , \parSpace_1(\bSigma; \gamma_n)]  \rightarrow 1$ as $n$ goes to infinity, it suffices to show that   $\chisquare(\prob_{\mathcal{H}}^n, \prob_{\bm{0}}^n) = o(1)$.
By calculation we have
\#\label{eq::info_chi_square}
& \chisquare(\prob_{\mathcal{H}}^n, \prob_{\bm{0}}^n)   = \Expect_{\prob^n_{\bm{0}}}\left\{\left[ \frac{\dd\prob_{\mathcal{H}}^n}{\dd\prob_{\bm{0}}^n}(Y, \bX)   - 1\right]^2\right\} = \Expect_{\prob^n_{\bm{0}}}\left \{\left[\frac{\dd\prob_{\mathcal{H}}^n}{\dd\prob_{\bm{0}}^n} (Y, \bX) \right]^2\right\} - 1 \notag  \\
& \quad = \frac{1}{|\mathcal{H}(s)|^2} \sum_{\vb_1, \vb_2 \in \mathcal{H}(s)} \Expect_{\prob^n_{\bm{0}}}\left[\frac{\dd\prob_{\vb_1}^n\dd\prob_{\vb_2}^n}{\dd\prob_{\bm{0}}^n\dd\prob_{\bm{0}}^n}(Y, \bX) \right] - 1 \notag\\
&  \quad = \frac{1}{|\mathcal{H}(s)|^2} \sum_{\vb_1, \vb_2 \in \mathcal{H}(s)} \biggl\{ \Expect_{\prob_{\bm{0}}}\left[\frac{\dd\prob_{\vb_1}\dd\prob_{\vb_2}}{\dd\prob_{\bm{0}}\dd\prob_{\bm{0}}}(Y, \bX) \right]\biggr\} ^n - 1 .  
\#
We utilize the following lemma to obtain an upper bound for the last term of  \eqref{eq::info_chi_square}. See \S \ref{proof::lem::h_function} for the proof.
\begin{lemma}\label{lem::h_function}
	For any  $\vb_1, \vb_2 \in \cH(s)$, we have
	\$
	\EE_{\PP_{\bm 0}} \left[ \frac{\ud \PP_{\vb_1}}{\ud \PP_{ \bm 0}} \frac{\ud \PP_{\vb_2}}{\ud \PP_{\bm 0}} ( Y , \bX) \right] =   \cosh   \left (      \la \vb_1, \vb_2   \ra /2 \right ) +  \alpha^2  \sinh  \left( \la \vb_1, \vb_2\ra /2  \right).
	\$
\end{lemma}

By Lemma \ref{lem::h_function}, we have 
\#\label{eq::use_lemma_chi_square}
&\chisquare(\prob_{\mathcal{H}}^n, \prob_{\bm{0}}^n)  \notag\\
&\quad =  \frac{1}{ |\mathcal{H}(s)|^2 } \sum_{\vb_1, \vb_2 \in \mathcal{H}(s)} \left[\cosh\left(1/2 \cdot \inner{\vb_1}{\vb_2}\right) + \alpha^2\sinh\left(1/2 \cdot \inner{\vb_1}{\vb_2}\right)\right]^n - 1.
\#
We define $\mathcal{C} := \{\mathcal{S} \subseteq [d] : |\mathcal{S}| = s\}$, and let $\mathbb{U}_{\mathcal{C}}$ be the uniform distribution over $\mathcal{C}$. Let  $\mathcal{S}_1, \mathcal{S}_2 \sim \mathbb{U}_{\mathcal{C}}$ be  two independent random sets. Then by \eqref{eq::use_lemma_chi_square}, we have
\[
\chisquare(\prob_{\mathcal{H}}^n, \prob_{\bm{0}}^n) = \Expect_{\mathcal{S}_1, \mathcal{S}_2} \left[\cosh( \beta^2/2 \cdot | \mathcal{S}_1\cap \mathcal{S}_2|) + \alpha^2\sinh( \beta^2/2 \cdot |\mathcal{S}_1\cap \mathcal{S}_2|)\right]^n - 1.
\]
We use the next lemma, proved in \S\ref{proof:lem:h_upper_bound}, to bound the above right-hand side. 
\begin{lemma} \label{lem:h_upper_bound}
	For any $x \geq 0$ and $v \in [0,1]$, we have
	\begin{equation} \label{eq:max}
	\cosh(x) + v\sinh(x) \leq \exp( 2v x) \vee \cosh(2x).
	\end{equation}
\end{lemma}	
Proceeding with this result and letting random variable $Z \sim |\mathcal{S}_1\cap \mathcal{S}_2|$, we have
\begin{align}
\chisquare(\prob_{\mathcal{H}}^n, \prob_{\bm{0}}^n) & \leq \Expect_{Z} \left[\exp( {\alpha^2\beta^2 Z}) \vee \cosh(\beta^2Z) \right]^n - 1 \notag\\
&  = \Expect_{Z} \left[\exp( n {\alpha^2\beta^2 Z  } )\vee \cosh(\beta^2Z)^n \right] - 1 \notag \\
&   = \Expect_{Z} \left\{\exp( n {\alpha^2\beta^2 Z  }) \vee \Expect_{U}\left[\exp( {\beta^2Z U})\right] \right\} - 1,\label{eq:tmp_upper_bound}
\end{align}
where in the last step, we introduce a random variable $U$ that is the summation of $n$ independent Rademacher random variables over $\{ -1, 1\}$. 
Then we have $\cosh(\beta^2Z)^n = \Expect_{U}[\exp ( {\beta^2Z U})]$. 
By   \eqref{eq:tmp_upper_bound}, we have
\#
\chisquare(\prob_{\mathcal{H}}^n, \prob_{\bm{0}}^n)& \leq \Expect_{Z} \Expect_{U}\left[\exp( n {\alpha^2\beta^2 Z })  \vee \exp( {\beta^2Z U} )\right] - 1  \notag\\
&  = \Expect_{U} \Expect_{Z} \left\{ \exp( n{\alpha^2\beta^2 } )\vee \exp ( {\beta^2U} )\right\}^Z - 1 \notag\\
& \leq \mathbb{E}_{U} \biggl\{     \sup_{\mathcal{S}_1 \in \mathcal{C}}   \mathbb{E}_{\mathcal{S}_2}\left[\exp(n {\alpha^2\beta^2  })  \vee \exp( {\beta^2U} )\right]^{|\mathcal{S}_1\cap \mathcal{S}_2|} \biggr\} - 1. \label{eq:tmp_upper_bound2}
\#
Now we turn to bound the expectation over $\mathcal{S}_2$ in \eqref{eq:tmp_upper_bound2}. For any fixed $\mathcal{S}_1$, we have
\[
|\mathcal{S}_1 \cap \mathcal{S}_2| = {\sum_{i \in \mathcal{S}_1} } V_i,
\]
where $V_i$ is binary random variable that indicates whether $i \in \mathcal{S}_2$. It is known that $V_1, \ldots, V_d$ are negative associated. Hence  we have
\#\label{eq::yet_a_temp}
\mathbb{E}_{\mathcal{S}_2}\left[\exp(n\alpha^2\beta^2 ) \vee \exp({\beta^2 U} )\right]^{|\mathcal{S}_1\cap \mathcal{S}_2|} & \leq \prod_{i \in \mathcal{S}_1} \mathbb{E}_{V_i}\left[ \exp( n {\alpha^2\beta^2  } )\vee \exp( {\beta^2 U} )\right]^{V_i} \notag \\
&= \left\{ 1 +  s/d \cdot \left[\exp ( n {\alpha^2\beta^2  })  \vee \exp( {\beta^2 U} )- 1\right] \right\} ^s.
\#
Plugging \eqref{eq::yet_a_temp} into \eqref{eq:tmp_upper_bound2} and expanding the polynomial term, we have
\begin{align*}
 \chisquare(\prob_{\mathcal{H}}^n, \prob_{\bm{0}}^n) &\leq \sum_{k=1}^s {s \choose k} \cdot (s/d)^k \cdot  \mathbb{E}_{U}\left[\exp ( n {\alpha^2\beta^2  })  \vee \exp( {\beta^2 U} )- 1\right] ^k \\
& = \sum_{k=1}^s {s \choose k}\cdot \left(s/d\right)^k \cdot \Big(\left[ \exp ( n {\alpha^2\beta^2  })- 1\right]^k\cdot \prob( U <  n \alpha^2 )  \\
&\quad \quad + \mathbb{E}_{U}  \left \{   \left[ \exp( {\beta^2U})  - 1 \right ]^k  \,\big\vert\, U \geq \alpha^2n  \right\}\cdot \prob( U \geq n \alpha^2) \Big), \\
& \leq T_1 + T_2,
\end{align*}
where $T_1$ and $T_2$ are defined as
\begin{align*}
& T_1 := \sum_{k=1}^s {s \choose k} \cdot \left(s/d\right)^k \cdot \left[\exp ( n {\alpha^2\beta^2 })  - 1\right]^k \\
& T_2 := \sum_{k=1}^s {s \choose k} \cdot \left(s/d\right)^k\cdot  \mathbb{E}_{U} \left \{  \left  [\exp( {\beta^2 U}) - 1 \right ]^k \vert U \geq 0  \right\} \cdot \prob( U \geq 0).
\end{align*}
It remains to bound $T_1$ and $T_2$ respectively.

\vskip .1in
\paragraph{Bounding $T_1$.}  Under condition $s \beta^2 =\gamma_n = o( 1/\alpha ^2 \cdot s\log d /  n)$, we have $\beta^2 = o( 1/\alpha^2 \cdot  \log d/n)$. Hence, for any small constant $C > 0$, we have $\beta^2 \leq C \cdot  1/ \alpha^2 \cdot  \log d/n $ when $n$ is sufficiently large. Note that we assume $s = o(d^{1/2 - \delta})$ for some fixed constant $\delta > 0$. Then we have
\begin{align*}
T_1 & \leq \sum_{k=1}^s {s \choose k} \cdot \left(s/d\right)^k\cdot \exp( {\alpha^2\beta^2nk} )\leq \sum_{k=1}^s \left [  s^2e  / (kd) \right]^k\cdot \exp( \alpha^2\beta^2nk) \\
& \leq \sum_{k=1}^s \left [  s^2e  / (kd)\right]^k \cdot \exp( {C k\log d} )= \sum_{k=1}^s ( s^2e/k \cdot d^{ C -1})^k \leq \sum_{k=1}^s(e/k \cdot d^{ C - 2\delta })^k,
\end{align*}
where the second step follows from the fact that ${s \choose k} \leq  (es /k  )^k$. Note that $C$ is chosen arbitrarily, hence we  can always choose $C \leq \delta$. It implies that  $e/ k \cdot d^{C -2\delta }  = o(1)$. We thus conclude $T_1 = o(1)$.

\vskip .1in
\paragraph{Bounding $T_2$.} For term $T_2$, we observe that
\begin{align*}
T_2 & \leq  \sum_{k=1}^s (e /k \cdot s^2 / d  )^k \cdot \mathbb{E}_{U} \left\{ \left[ \exp( {\beta^2 |U|} )- 1 \right] ^k \right\} \\
& \leq  \sum_{k=1}^s  (e /k \cdot s^2 / d  )^k  \cdot  \mathbb{E}_{U} \left[ (\beta^2 |U|)^k + \exp ({\beta^2k |U| }) \cdot \ind(\beta^2 |U| \geq 1)\right ]  \\
& \leq T_3 + T_4,
\end{align*}
where $T_3$ and $T_4$ are defined as
\$
&T_3 := \sum_{k=1}^s \mathbb{E}_{U} (e/ k \cdot s^2 \beta^2 /d \cdot  |U| )^k , \\
& T_4 := \sum_{k=1}^s (e /k \cdot s^2 / d  )^k  \cdot \mathbb{E}_{U} \left [ \exp ({\beta^2k |U| }) \cdot \ind(\beta^2 |U| \geq 1) \right ].
\$

Note that $U$ is summation of $n$ i.i.d. centered sub-Gaussian random variables $U_i$ each  with Orlicz $\psi_2$-norm equal to one. Therefore, $U$ is also centered sub-Gaussian random variable with $\norm{U}_{\psi_2} \leq C\sqrt{n}$ for some constant $C$. Thus it holds that 
\$
\mathbb{E} ( |U|^k  )  \leq (\sqrt{k}  \cdot \norm{U}_{\psi_2})^k \leq   (C\sqrt{nk}  )^k.
\$ Hence for term $T_3$, we have 
\[
T_3 \leq \sum_{k = 1}^s \left [C es^2\beta^2\sqrt{n}/ (\sqrt{k}d )\right]^k,
\]
Under the condition $s\beta^2 = o (\sqrt{s \log d/ n })$, we have
\[
C es^2\beta^2\sqrt{n }/ ( \sqrt{k}d ) = o  \left( s\sqrt{s\log d}/ d \right).
\]
Since $s = o(\sqrt{d})$,  we have $s\sqrt{s\log d}/ d = o(1)$, which implies $T_3 = o(1)$.

To obtain an upper bound for term $T_4$, we let $W = \beta^2U$. So $W$ is centered sub-Gaussian with Orlicz norm $c\beta^2\sqrt{n}$. Computing integral by parts, we have
\begin{equation} \label{eq:E_U}
\mathbb{E}_{U}  \left[ \exp ({\beta^2k |U| }) \cdot \ind(\beta^2 |U| \geq 1) \right] = e^k \cdot \prob ( |W| \geq 1) + \int_{w = 1}^{\infty} ke^{wk}  \cdot \prob ( |W| \geq w) \dd w.
\end{equation}
Using the property of sub-Gaussianity, we have $\prob[W \geq t] \leq C_1\exp[ {-C_2 t^2/(\beta^2\sqrt{n})^2}]$ for some absolute constants $C_1, C_2 > 0$. Proceeding with \eqref{eq:E_U} and using shorthand $\sigma = \beta^2\sqrt{n}$, we obtain
\begin{align*}
&\mathbb{E}_{U} \left [ \exp ({\beta^2k |U| }) \cdot \ind(\beta^2 |U| \geq 1) \right]  \leq C_1e^ke^{-C_2 /\sigma^2} + C_1k\int_{w =1}^{\infty}e^{wk}e^{-C_2w^2/\sigma^2}\dd w \\
& \quad = C_1e^ke^{-C_2 /\sigma^2} + C_1 ke^{k^2\sigma^2/(4C_2)} \int_{w = 1}^{\infty} e^{-\frac{C_2}{\sigma^2}(w - \frac{k\sigma^2}{2C_2})^2}\dd w \leq C_1e^k + C_3 ke^{k^2\sigma^2/(4C_2)}\sigma,
\end{align*}
where $C_3$ is a constant that depends on $C_1$ and  $C_2$. Thus we have 
\#\label{eq::yet_useless}
T_4 \leq  \underbrace{\sum_{k=1}^sC_1  \left [s^2e^2 / (kd)\right ]^k }_{T_5} + \underbrace{\sum_{k=1}^s C_3  \sigma k\left[  s^2e^2/ ( kd) \cdot \exp(k/4 \cdot \sigma^2/ C_2) \right]  ^k}_{T_6}.
\#
Note that $s^2/d = o(1)$, we thus have $T_5 = o(1)$. Under condition $s\beta^2 = o(\sqrt{s\log d/n})$, for any small constant $C> 0$, when $n$ is large enough, we have \$ \exp(k/4 \cdot \sigma^2/ C_2)\leq\exp( {C k\log d/s} )\leq \exp( {C \log d} ) \leq d^C.\$ Plugging \eqref{eq::yet_useless} into $T_6$ and using $s^2 = o(d^{1-2\delta})$, we have that each term in the summation is less that
\$
T_6 \leq  \sum_{k=1}^s	\sigma k \left [    e^2/ ( kd^{2\delta - C})\right ]^k \lesssim  \sum_{k=1}^s k \sqrt{ \log d/ s} \cdot  \left [e^2/ ( d^{2\delta - C})\right ] ^k .
\$
Since the constant $C$ is chosen arbitrarily, we have $T_6 = o(1)$. Accordingly, $T_4 = o(1)$ and $T_2 = o(1)$.

Finally, combining everything together, we have $\chisquare(\prob_{\mathcal{H}}^n, \prob_{\bm{0}}^n) = o(1)$, which completes the proof.

\subsection{Proof of Theorem \ref{thm:info_upper_bound}} \label{proof:thm:info_upper_bound}
We begin with some basic properties of sample sets $\{\wb_i\}_{i=1}^n$ and $\{\ub_i\}_{i=1}^n$. We introduce the random vector $\bW := \bX - \bX'$ to capture the distribution of samples $\{\wb_i\}_{i=1}^n$. Here $\bX$ follows the model given in \eqref{eq:model_1}-\eqref{eq:model_2}, and $\bX'$ is an independent copy of $\bX$. We note  that the marginal distribution of $\bX$ is given by $1/2 \cdot \Gaussian(\meanv_0, \bSigma) + 1/2 \cdot \Gaussian(\meanv_1, \bSigma)$. Thus   $\bW$ follows a mixture distribution
\begin{equation} \label{eq:W_dist}
\bW \sim 1/2 \cdot \Gaussian(\bm{0}, 2\bSigma)  + 1/4 \cdot \Gaussian(\meanv_1 - \meanv_0, 2\bSigma) +1/4 \cdot\Gaussian(\meanv_0 - \meanv_1, 2\bSigma).
\end{equation}
Moreover, conditioning on the observed label $Y$,  the distribution of $\bX$ is given by
\begin{align}
\bX | Y = 0 & ~ \sim~ (1 + \alpha) /2 \cdot \Gaussian(\meanv_0, \bSigma) + (1 - \alpha) /2 \cdot \Gaussian(\meanv_1, \bSigma), \label{eq:cond_dist_0}\\
\bX | Y = 1 & ~ \sim~(1 + \alpha) /2\cdot \Gaussian(\meanv_1, \bSigma) +  (1 - \alpha) /2\cdot \Gaussian(\meanv_0, \bSigma). \label{eq:cond_dist_1}
\end{align}
We introduce a random vector $\bU := \bX^{(1)} - \bX^{(0)}$ that corresponds to samples $\{\ub_i\}_{i=1}^n$. Here random vectors $\bX^{(0)}$ and $\bX^{(1)}$ are independent and have distributions given in \eqref{eq:cond_dist_0}, \eqref{eq:cond_dist_1}, respectively. The distribution of $\bU$ is given by
\begin{equation} \label{eq:U_dist}
\bU \sim  (1 + \alpha) ^2/4\cdot \Gaussian(\meanv_1 - \meanv_0, 2\bSigma) +  ( 1 - \alpha^2) /2 \cdot \Gaussian(\bm{0}, 2\bSigma) +  (1 - \alpha) ^2/4\cdot \Gaussian(\meanv_0 - \meanv_1, 2\bSigma).
\end{equation}

Now we turn to prove Theorem \ref{thm:info_upper_bound}. It suffices to prove this result by bounding type-I and type-II errors separately. In the end, we will show that
\[
 \sup_{\btheta \in \parSpace_0(\bSigma)}     \prob_{\btheta}^n ( \phi = 1)  \leq 4d^{-1} ~~ \text{and} ~~ \sup_{\btheta \in \parSpace_1(\bSigma; \gamma_n)}  \prob_{\btheta}^n ( \phi = 0)  \leq 16d^{-1}.
\]
\vskip .1in
\paragraph{Type-I error.} 
Under the  null hypothesis $\btheta \in \parSpace_0(\bSigma)$,  \eqref{eq:W_dist} and \eqref{eq:U_dist} reduce to
\[
\bW \sim \Gaussian(\bm{0}, 2\bSigma),~~\bU \sim \Gaussian(\bm{0}, 2\bSigma).
\]
To bound the type-I error of function $\phi_1$, we first note that
\[
\frac{1}{n}\sum_{i=1}^{ n }(\vb^{\top}\bSigma^{-1}\wb_i)^2 = \vb^{\top}\widehat{\bSigma}_{W}\vb,
\] 
where we let $\widehat{\bSigma}_W := 1/n \cdot  \sum_{i = 1}^{n} \bSigma^{-1}\wb_i\wb_i^{\top}\bSigma^{-1}$, i.e., an empirical covariance matrix of random vector $\bSigma^{-1}\bW \sim \Gaussian(\bm{0}, 2\bSigma^{-1})$. For any matrix $\Ab \in \real^{d\times d}$ and $\mathcal{S} \subseteq [d]$, we let $[\Ab]_{\mathcal{S}} \in \real^{|\mathcal{S}| \times |\mathcal{S} |}$ be  the submatrix of $\Ab$, which contains the entries with row and column indices in $\mathcal{S}$. By standard tail bound of Gaussian covariance estimation (see Lemma \ref{lem:Gaussian_cov}),   for any fixed $\mathcal{S} \in [d]$ with $|\mathcal{S}| = s$, and  any $\epsilon \in (0,1)$, when $n \geq Cs/\epsilon^2$ for some constant $C$, we have
\begin{equation} \label{eq:spectral_norm_bound}
\prob_{\btheta} ^n\left [\opnorm{(\widehat{\bSigma}_W  - 2\bSigma^{-1})_{\mathcal{S}}}{2} \geq 2\epsilon\opnorm{(\bSigma^{-1})_{\mathcal{S}}}{2}  \right ]\leq 2e^{-n}.
\end{equation}
Note that $\opnorm{(\bSigma^{-1})_{\mathcal{S}}}{2} \leq \opnorm{\bSigma^{-1}}{2}$ for all $\mathcal{S} \subseteq [d]$. By taking union bound over all subsets with size $s$ in $[d]$, we have
\$ 
&\prob_{\btheta}^n\left[   \sup_{\mathcal{S} \in [d], |\mathcal{S}| = s}  \opnorm{(\widehat{\bSigma}_W  - 2\bSigma^{-1})_{\mathcal{S}}}{2} \geq 2\epsilon\opnorm{\bSigma^{-1}}{2}\right] \leq 2{d \choose s}e^{-n} \\
& \quad \overset{(a)}{\leq} 2 \exp\left[  {-n + s\log(ed/s)} \right] \overset{(b)}{\leq} 2[s / (ed)]^s \leq 2d^{-1}.
\$
Here step $(a)$ follows from the fact that ${d \choose s} \leq (ed /s)^s$ and  step $(b)$ follows from the assumption that $n \geq 2s\log(ed/s)$. In the last step we use the fact that function $f(s) = (s/d)^s$ is monotonically decreasing for $s \in [1, d/e]$. We set $\epsilon = \sqrt{s\log(ed/s)/n}$. Under condition $n \geq 2s\log(ed/s)$, we have $\epsilon < 1$. Moreover, when $s \leq C'd$ for sufficiently small constant $C'$ that depends on $C$, we have $n \geq C s/\epsilon^2$. Therefore, such value of $\epsilon$   leads to \eqref{eq:spectral_norm_bound}. Thus we conclude that
\[
\prob_{\theta}^n\left[\frac{\vb^{\top}\widehat{\bSigma}_{W}\vb - 2\vb^{\top}\bSigma^{-1}\vb}{2\vb^{\top}\bSigma^{-1}\vb} \geq \sqrt{\frac{s\log(ed/s)}{n}} \cdot \frac{\opnorm{\bSigma^{-1}}{2}}{\vb^{\top}\bSigma^{-1}\vb}, \text{for all}~ \vb\in \mathcal{B}_2(s) \right] \leq 2d^{-1}
\]
Note that $\opnorm{\bSigma^{-1}}{2}/(\vb^{\top}\bSigma^{-1}\vb) \leq \opnorm{\bSigma^{-1}}{2}\opnorm{\bSigma}{2} = \kappa$.  Our choice of $\tau_1$ ensures the type-I error of $\phi_1$ does not exceed $2d^{-1}$.

Now we turn to analyze the performance of $\phi_2$. Recall that $\phi_1$ simply selects the coordinate of $\bar{\ub} := 1/n \cdot \sum_{i=1}^n \ub_i$ that has the largest magnitude (scaled with $\diag(\bSigma)^{-1/2}$) and compare it with $\tau_2$. It suffices to show all coordinates are well bounded around $0$ under null hypothesis. Denote the $j$-th coordinate of $\bar{\ub}$ by $\bar{u}_j$. Denote the $j$-th diagonal term of $\bSigma$ by $\sigma_j$. We have $\bar{u}_j \sim \Gaussian(0, 2\sigma_j/n)$. Recall that for standard normal random variable $X$, we have
\begin{equation} \label{eq:Gaussian_tail}
\prob  (|X| \geq t) \leq 2\exp( {-t^2/2}) ~~\text{for any}~~ t \geq 1.
\end{equation}
Using this property and taking union bound over $j \in [d]$, we have
\[
\prob _{\theta}^n \left (  \sup_{j \in [d]}   |\bar{u}_j|/ \sqrt{\sigma_j}\geq  {8\log d/ n} \right) \leq 2d \cdot \exp( {-2\log d}) = 2d^{-1}.
\]
Accordingly, our choice of $\tau_2$ can ensure type-I error of $\phi_2$ is controlled within $2d^{-1}$.

\vskip .1in
\paragraph{Type-II error.}   Under the alternative hypothesis   $\btheta \in \parSpace_1(\bSigma; \gamma_n)$. Note that  $\phi = 0$ if and only if $\phi_1 = 0$ and $\phi_2 = 0$. Thus, for any $\btheta \in \parSpace_1(\bSigma; \gamma_n)$, we have
\begin{equation} \label{eq:type_two_phi}
\prob_{\btheta}^n( \phi = 0 ) =  \prob_{\btheta}^n( \phi_1 = 0 \cap \phi_2 = 0)\leq \prob_{\btheta}^n( \phi_1 = 0) \wedge \prob_{\btheta}^n( \phi_2 = 0).
\end{equation}
We assume $\gamma_n \geq C\kappa [\sqrt{s\log d/n}\vee  (1/\alpha^2 \cdot s\log d/n)]$. It suffices to bound the  type-II error by considering these two cases: (i) when $\gamma_n \gtrsim  \kappa\sqrt{s\log d/n}$, we show that  $\prob_{\btheta}^n( \phi_1 = 0)  \leq 16d^{-1}$; (ii) when $\gamma_n \gtrsim \kappa/ \alpha^2\cdot   s\log d/n$ and $16/\alpha^2  \cdot s\log d/n \leq \sqrt{s\log d/n}$, we show $\prob_{\btheta}^n[\phi_2 = 0] \leq 7d^{-1}$.
\vskip .1in
\paragraph{Case (i).} Now we consider the first case. We denote $\Delta\bmu := \meanv_1 - \meanv_0$. Let $\vb^* := \Delta{\meanv}/\twonorm{\Delta{\meanv}}$. Since $\vb^* \in \mathcal{B}_2(s)$, we have
\[
\sup_{\vb \in \mathcal{B}_2(s)} \frac{\vb^{\top}\widehat{\bSigma}_{W}\vb}{2\vb^{\top}\bSigma^{-1}\vb} \geq \frac{\vb^{*\top}\widehat{\bSigma}_{W}\vb^*}{2\vb^{*\top}\bSigma^{-1}\vb^*}.
\]
It remains to show the right hand side is larger than $1 + \tau_1$ with high probability. Note that
\[
\vb^{*\top}\widehat{\bSigma}_{W}\vb^* = \frac{1}{n}\sum_{i=1}^n (\vb^{*\top}\bSigma^{-1}\wb_i)^2.
\]
We define a random variable $\widetilde{W} := \vb^{*\top}\bSigma^{-1}\bW$, whose probability distribution is given by  \begin{equation} \label{eq:new_w_dist}
1/2 \cdot \Gaussian(0, \nu) + 1/4 \cdot \Gaussian(m , \nu) +1/4 \cdot \Gaussian(-m, \nu),
\end{equation}
where  we define $m := \rho(\btheta)/\twonorm{\Delta{\meanv}}$ and $\nu := 2\rho(\btheta)/\twonorm{\Delta{\meanv}}^2$.
Recall that $\rho(\btheta) := \Delta{\meanv}^{\top}\bSigma^{-1}\Delta{\meanv}$. Let $\widetilde{w}_i := \vb^{*\top}\bSigma^{-1}\wb_i$. Due to the mixture structure \eqref{eq:new_w_dist}, we can thus cluster $\{\widetilde{w}_i\}_{i=1}^n$ into three groups $\{\widetilde{w}_i^{(k)}\}_{i=1}^{n_k}, k \in \{1,2,3\}$, based on the latent labels. The $k$-th group corresponds to the $k$-th term in \eqref{eq:new_w_dist}. Note that $\Expect(n_1) = n/2, \Expect(n_2) = \Expect(n_3)= n/4$. Define event $\mathcal{E}_1$ as
\begin{equation} \label{eq:E_1}
\mathcal{E}_1 := \left\{ \left| n_1 - n/2\right| \leq 1/8 \cdot n,~ \left| n_2 - n/4 \right| \leq 1/8 \cdot n,~ \left| n_3 - n/4\right| \leq 1/8 \cdot n\right\}.
\end{equation}
By Hoeffding's inequality, we have $\prob( \mathcal{E}_1)  \geq 1 - 6\exp( {-n^2/32})$. 

From now on, we condition on event $\mathcal{E}_1$. By the standard $\chi^2$-tail bound (Lemma \ref{lem:chi-square}),  for any  $t \in (0,1)$ and $k \in \{1,2,3\}$, we have
\begin{equation} \label{eq:w_square}
\prob_{\btheta}^n \left(\left| \sum_{i=1}^{n_k} (\widetilde{w}_i^{(k)} - m_k)^2 - n_k\nu\right| \geq n_k\nu t\right)  \leq 2e^{-n_kt^2/8} \leq 2e^{-nt^2/64},
\end{equation}
where $m_1 = 0, m_2 = -m_3 = m$. Moreover, using tail bound of Gaussian \eqref{eq:Gaussian_tail}, for $t' \geq 1/\sqrt{n_k}$ and $k = 2,3$,
\begin{equation} \label{eq:w_sum}
\prob_{\btheta}^n \left(\left|  \sum_{i=1}^{n_k}  \widetilde{w}_i^{(k)} - n_k m_k\right| \geq n_k \sqrt{\nu}t'\right) \leq 2e^{-n_kt'^2/2} \leq  2e^{-nt'^2/16}.
\end{equation}
Excluding the small chance events in \eqref{eq:w_square} and \eqref{eq:w_sum}, we find that
\begin{align*}
 \sum_{i=1}^n \widetilde{w}^2_i & = \sum_{k=1}^3 \sum_{i=1}^{n_k} (\widetilde{w}_i^{(k)} - m_k)^2 + 2\sum_{k=2}^{3} \sum_{i=1}^{n_k}m_k\widetilde{w}_i^{(k)} - (n_2 + n_3)m^2 \\
& \geq  n\nu(1-t) + 2\sum_{k=2}^{3} \sum_{i=1}^{n_k}m_k\widetilde{w}_i^{(k)} - (n_2 + n_3)m^2 \\
&  \geq  n\nu(1-t) + (n_2 + n_3)m^2 - 2(n_2+n_3)\sqrt{\nu}t'm \\
&    \geq  n\nu(1-t) + 1/4 \cdot  nm^2 - 3/2 \cdot  n\sqrt{\nu}t'm,
\end{align*}	
where the last step follows from \eqref{eq:E_1}.
Note that $2\vb^{*\top}\bSigma^{-1}\vb^* = \nu$. We thus have
\# \label{eq:phi_1_lower_bound}
\frac{ \vb^{^*\top}\widehat{\bSigma}_{W}\vb^*}{2   \vb^{*\top}\bSigma^{-1}\vb^*} - 1 & = \frac{  \sum_{i=1}^n  \widetilde{w}^2_i }{2n \vb^{*\top}\bSigma^{-1}\vb^* }- 1 \geq \frac{m^2}  {4\nu} - t -  \frac{3 m t'}{2\sqrt{\nu}}\notag\\
&  =1/8 \cdot \snr(\btheta) - t - 3t' /4 \cdot \sqrt{2\snr(\btheta)}.	
\#
Now we choose $t = t' = 8\sqrt{s\log(ed/s)/n}$, which is less than one under condition $n \geq 64s\log(ed/s)$. When $\rho(\btheta) \geq C\kappa \sqrt{s\log(ed/s)/n}$ for sufficiently large constant $C$, we can have $t \leq \rho(\btheta)/32$ and $t' \leq \sqrt{t'} \leq \sqrt{\rho(\btheta)}/48$. Accordingly, proceeding with \eqref{eq:phi_1_lower_bound} gives
\[
1/2 \cdot  \vb^{*\top}\widehat{\bSigma}_{W}\vb^*/\vb^{*\top}\bSigma^{-1}\vb^*  - 1\geq 1/16\cdot \rho(\btheta) \geq \tau_1.
\]
Plugging the value of $t, t'$ into the tail bounds in \eqref{eq:w_square} \eqref{eq:w_sum} and using the probability of event $\mathcal{E}_1$, we have the type-II error of $\phi_1$ is most $10d^{-1} + 6e^{-n^2/32} \leq 16d^{-1}$.

\vskip .1in
\paragraph{Case (ii).} Now we turn to analyze the performance of $\phi_2$. We introduce shorthands $\widetilde{\meanv} := \diag(\bSigma)^{-1/2}\Delta{\meanv}$ and $\bLambda := \diag(\bSigma)^{1/2}$. Then it holds that
\begin{align*}
\rho(\btheta) & = \Delta{\meanv}^{\top}\bSigma^{-1}\Delta{\meanv} = \Delta{\meanv}^{\top}\bLambda^{-1} \bLambda\bSigma^{-1}\bLambda\bLambda^{-1} \Delta{\meanv} \leq \twonorm{\widetilde{\meanv}}^2\opnorm{\bLambda\bSigma^{-1}\bLambda}{op} \\
& \leq \twonorm{\widetilde{\meanv}}^2 \opnorm{\bLambda}{2}^2 \opnorm{\Sigma^{-1}}{2} \leq \kappa \twonorm{\widetilde{\meanv}}^2,
\end{align*}
where the last step follows from the fact that $\opnorm{\diag(\bSigma)}{2} \leq \opnorm{\bSigma}{2}$.
Suppose the $j$-th coordinate of $\widetilde{\meanv}$, denoted by $\beta$, has largest magnitude. Since $\twonorm{\widetilde{\ub}}^2 \leq s \beta^2$, we have $\beta^2 \geq \rho(\btheta)/(s\kappa)$. Under condition \$\rho(\btheta) \geq \gamma_n \geq \frac{400\kappa s\log d}{\alpha^2n},\$ we have 
\begin{equation} \label{eq:beta_lower_bound}
\beta \geq 20\sqrt{\log d/(\alpha^2n)}.
\end{equation} 
Let $\vb^* = \sign(\beta)\cdot \eb_j$. We have
\[
\sup_{\vb \in \mathcal{B}_2(1)}  \inner{\vb}{\bLambda^{-1}\bar{\ub}} \geq \inner{\vb^*}{\bLambda^{-1}\bar{\ub}} = \left|\frac{1}{n} \sum_{i=1}^n \widetilde{u}_{ij}\right|,
\]
where we denote the $j$-th coordinate of $\bLambda^{-1}\ub_i$ by $\widetilde{u}_{ij}$. 

Let $U_j$ be the $j$-th coordinate of $\bU$. Note that $\{\widetilde{u}_{ij}\}_{i=1}^n$ are i.i.d. samples of $U_j/\sqrt{\sigma_j}$. Recall that $\sigma_j$ is the $j$-th diagonal term of $\bSigma$. According to \eqref{eq:U_dist}, $U_j/\sqrt{\sigma_j}$ has the mixture distribution
\begin{equation} \label{eq:U_j}
(1 + \alpha)^2/ 4 \cdot \Gaussian(\beta, 2) + (1 - \alpha^2 )/2\cdot \Gaussian(0, 2) + (1 - \alpha)^2/ 4 \cdot\Gaussian(-\beta, 2).
\end{equation}
We can cluster these samples into three groups $ \{\widetilde{u}_{ij}^{(k)}\}_{i=1}^{n_k}, k \in \{1,2,3\}$ based on latent labels, where $k$-th group corresponds to the $k$-th term in \eqref{eq:U_j}. Using tail bound of Gaussian \eqref{eq:Gaussian_tail}, we have for $t \geq 1$ and $k \in \{1,2,3\}$,
\[
\prob_{\btheta}^n\left( \left| \sum_{i=1}^{n_k } \widetilde{u}^{(k)}_{ij} - n_km_k\right| \geq \sqrt{2n_k}t\right) \leq 2e^{-t^2/2},
\]
where $m_1 = -m_3 = \beta, m_2 = 0$. Therefore, with probability at least $1 - 6e^{-t^2/2}$, it holds that 
\begin{equation} \label{eq:sum_u_ij}
\left| \frac{1}{n} \sum_{i=1}^n \widetilde{u}_{ij} - \frac{(n_1 - n_3) \beta}{n} \right| \leq  t \cdot \sum_{k = 1}^3 \sqrt{ \frac{2n_k}{  n^2} } \leq \frac{ 5t}{ \sqrt{n}}.
\end{equation}
It remains to bound $n_1 - n_3$.  Note that $n_1 - n_3$ is a summation of $n$ i.i.d. random variables $V_i$ satisfying $\prob( V_i = 1) = (1+\alpha)^2/4$,  $\prob(V_i = 0) = (1-\alpha^2)/2$, and  $\prob( V_i = -1) = (1-\alpha)^2/4$. Then $V_i$ has mean $\alpha$, variance $(1 - \alpha^2)/2 \leq 1-\alpha$, and $|V_i - \mathbb{E}(V_i)| \leq 2$. By Bernstein's inequality, we have that for $t' > 0$,
\[
\prob\left(\left|n_1 - n_3 - \alpha n\right| \geq t'\right) \leq \exp \left [ -\frac{ t'^2  }{2(1-\alpha)n +4t'/3 }\right ].
\]
Choosing $t' = \alpha n/2$, we thus have
\begin{equation} \label{eq:coordinate_bound_2}
\prob\left( \left|n_1 - n_3 - \alpha\cdot  n\right| \geq \alpha n/ 2 \right) \leq \exp \left [-\frac{  \alpha^2n} {8   (1-\alpha) +8 \alpha/3 } \right] \leq \exp( -\alpha^2 n/8) \leq d^{-1},
\end{equation}
where the last step follows from condition $8 s\log d/(\alpha^2n) \leq \sqrt{s\log(ed/s)/n} \leq 1$. Combining \eqref{eq:sum_u_ij} and \eqref{eq:coordinate_bound_2}, we have that with high probability $1 - 6e^{-t^2/2} - d^{-1}$,
\[
\left|1/n \cdot {\sum_{i=1}^n } \widetilde{u}_{ij}\right| \geq \alpha \beta/2 - 5t/ \sqrt{n}\geq 10\sqrt{\log d/ n}- {5t/\sqrt{n} }\geq \tau_2,
\]
where the second step follows from \eqref{eq:beta_lower_bound} and the last inequality holds by setting $ t = \sqrt{2\log d}$, which gives the type-II error of $\phi_2$ is at most $7d^{-1}$.

\vskip .1in
Using \eqref{eq:type_two_phi} and the conclusions in the above two cases, we thus show Type-II error of $\phi$ is at most $16d^{-1}$ and thus complete the proof.

\subsection{Proof of Theorem \ref{thm::comp_limit}} \label{proof::thm::comp_limit}
In this section, we prove the computational lower bound.  We first show that the information-theoretic lower bound in \eqref{eq::main_result1} is a lower bound of the computationally tractable minimax rate. To see this, we consider the oracle $r^*$ that returns sample average $ n^{-1} \sum_{i=1}^n q (y_i, \xb_i)$ for any query function $q$. As discussed in \S \ref{sec::oracle_model}, Bernstein's inequality in \eqref{eq::bernstein} and uniform concentration of empirical process imply that $r^* \in \cR[\xi,n,T_n, \eta(\cQ_{\mA}) ]$. In addition, every test function $\phi  $ that is based on the responses of $r^*$ is also a function of $\{ (y_i, \xb_i )\}_{i=1}^n $.   Thus combining \eqref{eq::minimax_risk} and  \eqref{eq::minimax_risk_oracle},  it holds that 
\$
\overline{R}_n^*(\cG_0, \cG_1;\mA, r^*) \geq  \risk^*_n(\parSpace_0, \parSpace_1).
\$
Therefore, by Theorem \ref{thm:info_lower_bound}, for any $\gamma_n$ satisfying 
\$
\gamma_n = o \left[ \sqrt{ s\log d / n} \wedge   (  1/ \alpha^2  \cdot s\log d/   n) \right ],
\$
we have $\lim_{n \rightarrow \infty} \overline{R}_n^*[ \cG_0, \cG_1 (\gamma_n) ;\mA, r^* ] = 1$. Here the equality  holds because   a test based on purely random guess incurs risk one.

Based on this observation, to show Theorem \ref{thm::comp_limit}, it the following, we assume that 
\#\label{eq:signal_strength}
\gamma_n   = o \left [ \sqrt{ s^2 / n} \wedge   (  1/ \alpha^2  \cdot s/   n)  \right].
\# 
We show that under this assumption, there exists an oracle $r   $ such that the minimax testing risk is not negligible.
Similar to the derivation of the information theoretical lower bound, we also  focus on the restricted testing problem  defined in \eqref{eq::restricted_test}. Following the same notations, we denote by $\prob_{\bm{0}}$  the distribution of model   $  (\bm{0}, \bm{0}, \Ib, \alpha) $ and by $\prob _{\vb}$ the distribution of model  $(-\vb/2, \vb/2, \Ib, \alpha)$ for all $\vb \in \cH(s)= \{\ub \in \{0,\beta\}^d  \colon\zeronorm{\ub} = s\}$. Here we assume that the SNR under $H_1$ satisfies $\beta ^2 s  = \gamma_n$.

Moreover, we define $\overline{\PP}_{\bm{0}} $ as the distribution of the random variables returned by the  statistical query model  under the null hypothesis $H_0 $ and define $\overline{\PP}_{\vb}$   correspondingly.
Then  the minimax testing risk $\overline {R}_n^* ( \cG_0, \cG_1; \mA, r)$ defined in \eqref{eq::minimax_risk_oracle}  is lower bounded by
\$
  \sup_{\bSigma} \overline{R}_n^* [ \cG_0(\bSigma), \cG_1(\bSigma;\gamma_n); \mA , r ]  \geq   \inf_{\phi\in \cH(\mA,r)}     \left[\overline {\prob}_{\bm{0}} (\phi = 1) + \frac{1}{ |\mathcal{H}(s)|} \sum_{\vb \in \mathcal{H}(s) } \overline{\prob}_{\vb}( \phi = 0)\right ].
\$

The following  lemma establishes a sufficient condition that any hypothesis test  under  the statistical query model   is asymptotically powerless. See \cite{wang2015sharp} and \cite{fan2016curse} for a proof.

\begin{lemma} \label{lemma::distinguish}
	For any algorithm $\mA \in \cA(T)$ and any query function $q \in \cQ_{\mA}$, we define 	\$ \cC_1(q) &=   \left\{ \vb\in \cH (s): \EE_{\PP_\vb} \left [q(Y , \bX) \right] - \EE_{\PP_{\bm 0}} \left [q(Y , \bX) \right] >   \tau_{q}(\PP_{\vb})   \right\},  \\
 \cC_2 (q) &=   \left\{ \vb\in \cH (s):   \EE_{\PP_{\bm 0}} \left [q(Y , \bX) \right] - \EE_{\PP_\vb} \left [q(Y , \bX) \right]  > \tau_{q}(\PP_{\vb})    \right\}.
\$
	Here $\tau_{q}(\PP_{\vb}) $  is the tolerance parameter defined in \eqref{eq::query_2}  when $(Y, \bX) \sim \PP_{\vb}$. 
	Then if $T \cdot \sup _{q\in \cQ_{\mA}} \left(| \cC_1 (q) | + | \cC_2(q) | \right)/  | \cH(s)| = o(1)$,  there exists an oracle $r\in \cR[\xi,n,T, \eta(\cQ_{\mA}) ]$ such~that
	\$
	 \inf_{\phi\in \cH(\mA,r)}    \left[ \overline{\PP}_{\bm 0}(\phi = 1) +  \frac{ 1}{ |\mathcal{H}(s)| }   \sum_{\vb \in \mathcal{H}(s) }\overline{\PP}_{\vb}(\phi = 0) \right ]  = 1.
	\$
	\end{lemma}
	By this lemma, we need to construct an upper bound for  $  \sup _{q\in \cQ_{\mA}} \left(| \cC_1 (q) | + | \cC_2(q) | \right)$. In the sequel, we achieve this goal by studying the uniform mixture of $\{ \PP_{\vb} \colon \vb \in \cC_{\ell}(q)\}$ for $\ell \in \{1,2\}$. Specifically, we define
\#\label{eq::define_two_mixtures}
\PP_{ \cC_1(q)} =  
\frac{1}{ |\cC_1(q)| }   \sum_{ \vb \in \cC_1(q)}      \PP_{\vb}~~\text{and}~~\PP_{ \cC_2(q)} = \frac{1}{ |\cC_2(q)|} \sum_{\vb \in \cC_2(q)}  \PP_{\vb}.
\#
The following lemma, obtained from \cite{fan2016curse}, establishes an upper bound for the $\chi^2$-divergence between $\PP_{\cC_{\ell}(q)}$ and $\PP_{\bm 0}$.
	\begin{lemma}\label{lemma::distinguish2}
For  $\ell\in \{1, 2\}$   we define 
\#\label{def::bar_C}
\overline{ \cC}_{\ell} (q, \vb)   = \argmax_{ {\cC}}  \Biggl \{  \frac{1}{|  { \cC}|}  {  \sum_{\vb'\in {\cC \subseteq \cH(s) }  } } \EE_{\PP_{\bm 0}}  \left[\frac{\ud \PP_{\vb}}{\ud \PP_{\bm 0}} \frac{\ud \PP_{\vb'}} {\ud \PP_{\bm 0}}( Y , \bX) \right]-1~ \bigg \vert  ~|  {\cC} | = | \cC_\ell(q) |\Biggr\}.  
\# 
Then the   $\chi^2$-divergence between $\PP_{\cC_{\ell(q)}}$ and $\PP_{\bm 0}$ is bounded by
\#\label{eq::chi_square_div2}
D_{\chi^2} (\PP_{ \cC_{\ell}(q)}, \PP_{\bm 0} )\leq \sup_{\vb\in \cC_{\ell}(q) }  \frac{1}{|\cC_{\ell}(q)|} {\sum_{\vb'\in \overline{\cC }_{\ell} (q, \vb ) } } \EE_{\PP_{\bm 0}}  \left[\frac{\ud \PP_{\vb }}{\ud \PP_{\bm 0}} \frac{\ud \PP_{\vb'}}{\ud \PP_{\bm 0}}( Y , \bX) \right]- 1 .
\#
\end{lemma}

Notice that Lemma \ref{lem::h_function}  enables us to compute the right-hand side of \eqref{eq::chi_square_div2} in closed form.
For any $\alpha \in [0, 1]$,  function $h_{\alpha} (t) = \cosh[\beta^2 /2 \cdot (s - t) ] + \alpha^2 \sinh[\beta^2/ 2 \cdot (s - t)]$  is monotone nonincreasing for $t \in \{0, \ldots, s\}$ and $f(s) = 0$.~In addition, for any  $\vb\in \cH(s)$ and any $j \in\{0,\ldots, s\}$, we define
\#\label{eq::def_cj}
\cC_j(\vb) = \left \{\vb' \in \cH (s) : |\supp(\vb)  \cap \supp(\vb')| = s-j\right\}.
\#
For $\ell \in \{ 1,2 \}$, any  query function $q \in \cQ_{\mA}$, and any~$\vb \in \cC_{\ell} (q)$, by Lemma \ref{lem::h_function} and  the definition of~$\overline{\cC}_{\ell}(q, \vb)$ in~\eqref{def::bar_C},  there exists  an integer  $k_{\ell} (q, \vb )$  that satisfies
\#\label{eq::def_kq1}
\overline{\cC}_{\ell}(q, \vb) = \cC_{0}(\vb) \cup \cC_1(\vb) \cup \cdots \cup  \cC_{k_{\ell} (q, \vb) -1}(\vb) \cup \cC'_{\ell} (q,\vb ),
\#
where $\cC'_{\ell} (q,\vb) = \overline \cC_{\ell}(q, \vb ) \setminus {\textstyle\bigcup_{j=0}^{k_{\ell}(q, \vb) - 1}} \cC_j(\vb)$ has cardinality
\#\label{eq::def_kq2}
|\cC'_{\ell} (q,\vb)| = | \cC_{\ell}(q)| -   \sum_{  j = 0}^{ k_{\ell}(q,\vb) -1}  | \cC_{j} (\vb)| < | \cC_{k_{\ell} (q, \vb )} (\vb)|.
\#
Thus we can sandwich the cardinality of $\overline{\cC} _{\ell} (q , \vb)$ by
\#\label{eq::set_bound}
 \sum_{j=0}^{k_{\ell}(q, \vb ) }  |\cC_j(\vb) | > | \overline{ \cC} _{\ell} (q , \vb)|\geq  \sum_{j=0}^{k_{\ell} (q, \vb )-1} |\cC_j (\vb) |.
\#
Combining   Lemmas  \ref{lem::h_function} and \ref{lemma::distinguish2}, we further have
\#\label{eq::upper_bound_chi_square1}
1 + D_{\chi^2} (\PP_{  \cC_{\ell}(q)}, \PP_{\bm 0}   )\leq \frac{\sum_{i=0}^ {k_{\ell}(q, \vb )-1} h_{\alpha}(j)\cdot |\cC_j(\vb)| + h_{\alpha}[k_{\ell}(q, \vb )] \cdot | \cC_{\ell}'(q, \vb ) |}  { \sum _{j=0}^{k_{\ell}(q, \vb)-1} | \cC_j(\vb)| + | \cC_{\ell}'(q, \vb)|},~\text{for all}~\vb \in \cC_{\ell} (q).
\#
Moreover, by \eqref{eq::upper_bound_chi_square1} and~the   monotonicity of $h_{\alpha}(t)$  we obtain
\#\label{eq::upper_bound_chi_square}
1 + D_{\chi^2} (\PP_{ \cC_{\ell}(q)}, \PP_{\bm 0} )  \leq \frac{\sum_{i=0}^ {k_{\ell}(q , \vb )-1} h_{\alpha}(j) \cdot    | \cC_j(\vb)| }  { \sum _{j=0}^{k_{\ell}(q, \vb )-1} | \cC_j(\vb)| }.
\#

By the definition of $\cC_j(\vb)$ in \eqref{eq::def_cj}, the cardinality of $\cC_j(\vb)$  does not depend on the choice of $\vb \in \cH(s)$ and we have $| \cC_j(\vb) | = {s \choose s - j}{d-s \choose j}$. Thus  for any  $j \in \{ 0, \ldots, s -1 \}$  we have 
\#\label{eq::compare_geometric}
| \cC_{j+1}(\vb) | / | \cC_j(\vb)| = (s- j) \cdot (d-s-j) / (j+1)^2 \geq (d-2s)/ s^2.
\#
Under the assumption that $s^2 / d = o(1)$, the right-hand side of \eqref{eq::compare_geometric} is lower bounded by $\zeta = d / (2s^2)$ when $d$ and $s$ are sufficiently large. 
Then we have  $| \cC_{j}(\vb)|  \leq \zeta^{j-s} | \cC_{s}(\vb)| $ for $  j \in \{0,\ldots, s\}$.
By the definition of $k_{\ell}(q, \vb )$ in \eqref{eq::def_kq1} and \eqref{eq::def_kq2}, for any  $q \in \cQ_{\mA}$, we further obtain
\#\label{eq::upper_bound_Cq}
|\cC_{\ell}(q)| & \leq \sum_{j=0}^{k_{\ell}(q, \vb ) } |\cC_j(\vb)|
\leq   |\cC_{ s }(\vb)|   \sum_{j=0}^{k_{\ell}(q, \vb ) } \zeta^{j - s} \notag \\
 &\leq \frac{\zeta^{-[ s  - k_{\ell}(q, \vb ) ]} |\cH(s) |}{1 - \zeta^{-1}} \leq 2\zeta^{-[ s  - k_{\ell}(q, \vb ) ] } | \cH(s)| ,
\#
where the last inequality follows from the fact that   $\zeta^{-1} = 2s^2 / d= o(1)$.

Moreover, for any two positive sequences $\{ a_i \}_{i=0}^s$ and $\{ b_i \}_{i=0}^s$ satisfying $a_{i}/ a_{i-1} \geq b_{i } / b_{i-1} >1$ for all $i \in [s]$, since $h_{\alpha}(t)$ is nonincreasing, for any $k \in [s]$, we have 
\#\label{eq::useless}
\sum_{0\leq i< j \leq k} (a_i b_j - a_j b_i ) \cdot [ h_{\alpha} (i) - h_{\alpha} (j) ] \leq 0.
\#
Further   simplifying the terms in \eqref{eq::useless}, we have
\#\label{eq::useless2}
\frac{ \sum_{i=0}^k  [a_i h_{\alpha}(i)]}  {  \sum_{i=0}^k a_i  } \leq \frac{ \sum_{i = 0}^k  [b_i h_{\alpha} (i)]}  { \sum_{i=0}^k  b_i}.
\#

In what follows, we   upper bound  $k_{\ell}(q,\vb)$ for $\ell \in \{1,2\}$ and $\vb \in \cC_{\ell}(q)$. We employ the  shorthand $k_{\ell} = k_{\ell}(q, \vb)$ to simplify the notations.
Combining \eqref{eq::chi_square_div2}, \eqref{eq::upper_bound_chi_square}, and \eqref{eq::useless2} with   $a_j = | \cC_j(\vb)|$~and $b_j = \zeta^j$, we  have
\#\label{eq::bound_divergence_new}
& 1 + D_{\chi^2} (\PP_{ \cC_{\ell}(q)}, \PP_{\bm 0} )\leq \frac{ \sum_{j=0}^{k_{\ell}-1} \zeta^j h_{\alpha}(j) }{\sum_{j=0}^{k_{\ell}-1} \zeta^j} \notag \\
&\quad  =    \frac{ \sum_{j=0}^{k_{\ell}-1} \zeta^j  \left \{ \cosh\left[ \beta^2 / 2 \cdot (s- j)\right]  + \alpha^2  \sinh\left[ \beta^2 /2 \cdot (s-j) \right]\right\} } {\sum_{j=0}^{k_{\ell}-1} \zeta^j} \notag\\
&\quad \leq      \frac{ \sum_{j=0}^{k_{\ell}-1} \zeta^j  \cdot \left \{  \cosh\left[ \beta^2    (s- j)\right] \vee\exp\left[ \alpha^2   \beta^2  (s-j) \right] \right \}     }{\sum_{j=0}^{k_{\ell}-1} \zeta^j}    
\#
Here the second inequality follows from  Lemma \ref{lem:h_upper_bound}.  
For notational simplicity, we denote for  any $t \in\{0,\ldots, s\}$, we define
\$
f(t) = \cosh \left[\beta^2   (s-t) \right], ~~g(t) =  \exp\left[  \alpha^2 \beta^2   (s-t) \right].
\$
Note that both  $h(t)$  and $g(t)$ are monotone non-increasing, and thus $f(t) \geq f(s) = 1$ and $g(t) \geq g(s) = 1$. Moreover, by calculation, we have 
$$ f(j- 1) / f(j) \geq \cosh( \beta^2  ) \quad \text{and}\quad g(j-1) / g(j) = \exp(\alpha ^2 \beta^2 )$$ for all $j \in \{ 1,\ldots, s\}$. 
Thus, for all  $j \in  \{0, \ldots, k_{\ell}-1\}$, we have 
\# \label{eq:bound_fg}
f(j ) \leq f(k_{\ell} - 1) \cdot \left[ \cosh (\beta^2) \right]^{ k_{\ell}  - j - 1}   \quad \text{and}\quad g(j ) = g(k_{\ell} - 1)   \cdot \left[ \exp( \alpha ^2 \beta^2) \right]^{ k_{\ell}  - j - 1} .
\#
Besides, we denote $ \cosh( \beta^2  ) \vee  \exp(\alpha ^2 \beta^2 )$ by $\varphi(\beta)$ hereafter for simplicity.
By \eqref{eq:bound_fg}, we can further bound the last term  in  \eqref{eq::bound_divergence_new} by  
\#\label{eq::upper_sub1}
& \frac{ \sum_{j=0}^{k_{\ell}-1} \zeta^j \left \{  f(j ) \vee g(j) \right \}   }{\sum_{j=0}^{k_{\ell}-1} \zeta^j}  \leq  \bigl\{ f(k_{\ell} -1) \vee g(k_{\ell} -1) \bigr \} \cdot  \frac{ \sum_{j=0}^{k_{\ell}-1} \zeta^j   \left[ \varphi(\beta)\right]^{k_{\ell} - j +1}   }{\sum_{j=0}^{k_{\ell}-1} \zeta^j}    \notag\\
& \quad \leq \bigl\{ f(k_{\ell} -1) \vee g(k_{\ell} -1) \bigr \} \cdot   \frac{ \sum_{j=0}^{k_{\ell}-1}    \left[ \varphi(\beta) / \zeta\right]^{k_{\ell} - j +1}   }{\sum_{j=0}^{k_{\ell}-1} \zeta^{ -( k_{\ell}  - j +1  ) } } \notag \\
& \quad = \bigl\{ f(k_{\ell} -1) \vee g(k_{\ell} -1) \bigr \}\cdot \frac{1  - \left[\varphi(\beta)/ \zeta \right]^{k_{\ell} }}{ 1 - \zeta^{-k_{\ell}} } \cdot \frac{ 1- \zeta^{-1}} { 1 - \zeta^{-1} \varphi(\beta) } .
\#	
Note that $\varphi(\beta) > 1$ by definition. Thus,  it holds that 
\#\label{eq::upper_sub2}
\frac{1  - \left[\varphi(\beta)/ \zeta \right]^{k_{\ell} }}{ 1 - \zeta^{-k_{\ell}} } \leq 1 .
\#
Therefore, combining \eqref{eq::bound_divergence_new}, \eqref{eq::upper_sub1}, and  \eqref{eq::upper_sub2}, we obtain that  
\#\label{eq::final_bound_divergence}
& 1 + D_{\chi^2} (\PP_{ \cC_{\ell}(q)}, \PP_{\bm 0} )  \notag \\
& \quad \leq  \frac{  1- \zeta^{-1}     }  { 1 - \zeta^{-1} \varphi (\beta ) }  \cdot  \left\{ \cosh\left[ \beta^2 (s - k_{\ell} + 1)\right]    \bigvee   \exp\left[ \alpha^2 \beta^2 (s - k_{\ell} + 1)\right]    \right\} .
\#
Moreover,  we use the  following lemma  obtained from \cite{wang2015sharp} to establish a lower bound for $D_{\chi^2} (\PP_{ \cC_{\ell}(q)}, \PP_{\bm 0} )$.
\begin{lemma}\label{lem:1}\label{lemma::bound_chi_square_div}
	For any query function $q$ and $\ell\in \{1,2\}$, we have
	\$
	D_{\chi^2} (\PP_{  \cC_\ell(q)}, \PP_{\bm 0} ) \geq \log (T/\xi)/n.
	\$
\end{lemma}
We denote $\sqrt{\log (T/\xi)/n}$ by $\tau$  for simplicity of notations. Combining  \eqref{eq::final_bound_divergence},  Lemma \ref{lemma::bound_chi_square_div} and inequality $\cosh(x) \leq \exp(x^2/2)$, at least one of the two inequality holds
\#
(1+  \tau^2 ) \cdot \left[{1 - \zeta^{-1}\cdot \varphi(\beta)} \right]/ (1- \zeta^{-1} )   &\leq 
\exp \left[ \beta^4 /2  \cdot (s - k_{\ell} + 1)^2 \right],\label{eq::final_bound21}\\
(1+  \tau^2 ) \cdot \left[{1 - \zeta^{-1}\cdot \varphi(\beta)} \right]/ (1- \zeta^{-1} )   &\leq 
\exp \left[ \alpha^2 \beta^2  (s - k_{\ell} + 1)  \right].\label{eq::final_bound22}
\#
If \eqref{eq::final_bound21} holds,   taking the logarithm of the both sides, we have 
\#\label{eq::final_bound31}
\beta^4 /2 \cdot ( s - k_{\ell} +1 ) ^2 \geq \log (1+ \tau^2) -\log \left[\frac{1-\zeta^{-1}}{1 - \zeta^{-1}\varphi(\beta)} \right] .
\#
Whereas if  \eqref{eq::final_bound22} is true, it holds that 
\#\label{eq::final_bound32}
\alpha^2   \beta^2 ( s - k_{\ell} +1 )  \geq \log (1+ \tau^2) -\log \left[\frac{1-\zeta^{-1}}{1 - \zeta^{-1}\cdot \varphi(\beta)} \right] .
\#
In addition, by  the fact that $[\cosh(\beta^2/2) +  \exp(\alpha^2 \beta^2)] /\zeta = o(1)$, we have
\#
& \log \left[\frac{1-\zeta^{-1}}{1 - \zeta^{-1} \cdot \varphi(\beta)} \right]  = \log \biggl \{ 1 + \frac{\zeta^{-1} \left[ \varphi(\beta)-1 \right]   }{1-  \zeta^{-1}\varphi(\beta) } \biggr\} \notag \\
&\quad  = \cO \Bigl \{  \zeta^{-1} \cdot \left[ \varphi(\beta)-1 \right] \Bigr  \} = \cO \bigl [  \zeta^{-1} \cdot (\alpha^2 \beta^2) \vee \beta^4\bigr ] , \label{eq::bound_by_taylor1} 
\#
where the last equality follows from the Taylor expansions of $\exp(x)$ and $\cosh(x)$. 
Since $\gamma_n = s\beta^2$, by \eqref{eq:signal_strength} we have $(\alpha^2 \beta^2) \vee \beta^4 = o( \log d/n)$.  Hence, by \eqref{eq::bound_by_taylor1},  the second terms on the right-hand sides of \eqref{eq::final_bound31} and \eqref{eq::final_bound32} are asymptotically negligible compared with $\log(1+ \tau^2)$. 
Therefore, by  \eqref{eq::final_bound31} and \eqref{eq::final_bound32}, for $\ell \in \{1,2\}$, at least one of the following two arguments hold:
\$
k_{\ell}(q, \vb) \leq  s+1 - \sqrt{\log (1+ \tau^2) / \beta^4 } , ~~k_{\ell}(q, \vb) \leq      s + 1 - \log (1+ \tau^2) / (2\alpha^2 \beta^2).
\$
Equivalently, we have
\#\label{eq::final_bound4}
k_{\ell}(q, \vb) \leq \left  [  s+1 -\sqrt{\log (1+ \tau^2) / \beta^4 } \right ] \vee  \left [ s+1 - \log (1+ \tau^2) / (2\alpha^2 \beta^2)\right].
\#
Recall that $\tau = \sqrt{\log (T/\xi) / n} $  where $\xi = o(1)$.~For any constant $\eta >0$, we set $T = O(d^\eta)$.
By combining   Lemmas~\ref{lemma::distinguish} and \ref{lemma::distinguish2},  \eqref{eq::upper_bound_Cq},  and \eqref{eq::final_bound4}, we further obtain
\#\label{eq::bound_Main_term}
T \cdot   \frac{\sup  _{q\in \cQ_{\mA}} \left ( | \cC_1(q)| + | \cC_2(q) | \right ) }{ |\cH(s)| } & \leq 4T \cdot   \exp  \left  \{ -\log \zeta\cdot \left[\sqrt{  \log (1+\tau^2 ) / {\beta}^4 }- 1\right] \right \} \wedge\notag \\
& \quad \quad   4T \cdot   \exp  \left \{ -\log \zeta\cdot \left[   \log (1+\tau^2 ) /(2\alpha^2 \beta^2)- 1\right] \right\}  .
\#
Under the assumption of the theorem, there is a  sufficiently small  constant $\delta >0$ such that $s ^2 / d^{1- \delta } = O(1)$. Thus we have $\zeta = d/ (2s^2) =  \Omega(d^{\delta} )$. By inequality $\log(1+x) \geq x/2$, it holds that
$\log (1+ \tau^2) \geq \tau^2/2 = \log (T/\xi) / (2n) $. Under the condition  in \eqref{eq:signal_strength}
,~we have  \# \label{eq:compare_Main_term}
\frac{ \log (T/\xi) }{ 2 n\beta^4} \bigvee \frac{  \log (T/\xi)}{4 n\alpha^2 \beta^2}>\frac{\log (1/ \xi)}{ 2 n \beta^4}   \bigvee \frac{\log (1/ \xi)}{ 4 \alpha^2\beta^2}  \rightarrow \infty.\#
Hence if $n$ is sufficiently large, the left-hand side in \eqref{eq:compare_Main_term} is greater than an absolute constant $C$ 
 satisfiying $\delta (C-1) > \eta$.~Then by  \eqref{eq::bound_Main_term} we have
\#\label{eq::bound_key_quantity_1}
T \cdot   \frac{\sup  _{q\in \cQ_{\mA}} \left ( | \cC_1(q)| + | \cC_2(q) | \right ) }{ |\cH(s)| } & =  O [4 d^\eta \zeta^{-(C-1) }  ]=   O [4d^{\eta} d^{-\delta (C-1)} ] = o(1).
\#

Combining \eqref{eq::bound_key_quantity_1} and Lemma \ref{lemma::distinguish}, we conclude that $\overline {R}_n^* ( \cG_0, \cG_1; \mA, r) \rightarrow 1$ if \eqref{eq:signal_strength} holds.
This concludes the proof of Theorem \ref{thm::comp_limit}.

\subsection{Proof of Theorem \ref{thm::comp_upper} }\label{proof::thm::comp_upper}
To ease notation, we denote the joint distribution of $(Y, \bX)$ by $\PP_{\btheta}$ where the model parameter is given by $\btheta = ( \bmu_0, \bmu_1, \bSigma, \alpha)$. In addition, we let $\Delta \bmu = \bmu_1 - \bmu_0$. Thus $\Delta \bmu = {\bf 0}$ for all $\btheta \in \cG_0(\bSigma)$ and $\Delta \bmu \in \cB(s)$ for all $\btheta \in \cG_1(\bSigma;\gamma_n)$.
In what follows, we bound the type-I and type-II errors of $\overline{\phi} $ respectively.

\paragraph{Type-I error.}  For any $\btheta \in \cG_0(\bSigma)$, by the definition of $\overline{\phi}$, the type-I error is bounded by 
\$
\overline{\PP}_{\btheta} ( \overline \phi = 1) \leq \overline{\PP}_{\btheta} ( \overline{\phi}_1 = 1) + \overline{\PP}_{\btheta} ( \overline{\phi}_2 = 1).
\$
For test function $\overline{\phi}_1$,  since  marginally,   $\bX\sim 1/2\cdot  \Gaussian (\bmu_0, \bSigma) + 1/2 \cdot \Gaussian (\bmu_1,\bSigma)$, for any $\btheta \in \cG_0 (\bSigma) \cup \cG_1(\bSigma; \gamma_n)$, for any $j\in [d]$, we have 
\#
& \EE _{\PP_{\btheta} } ( X_j^2 / \sigma_j - 1 ) - \bigl[ \EE_{\PP_{\btheta}}  (X_j /\sqrt{\sigma_j} )   \bigr ] ^2  \notag \\
&\quad   = 1/4 \cdot ( \mu_{0,j} - \mu_{1, j})^2 / \sigma_{j} =    1/4 \cdot ( \Delta \bmu)_j^2 / \sigma_j,\label{eq::expected_query1}
\#
Here $\mu_{0,j}$ and $\mu_{1,j}$ denote the $j$-th entries of $\bmu_0$ and $\bmu_1$, and  $( \Delta \bmu)_j$ is the $j$-th entry of $\Delta \bmu$. 
In addition, by the definition of $q_j$ in \eqref{eq::query1} we have
\$
& \Bigl |\bigl  [\EE_{\PP_{\btheta}} q_j (Y, \bX)\bigr  ] ^2 -  \bigl[ \EE_{\PP_{\btheta}}  (X_j /\sqrt{\sigma_j} )   \bigr ] ^2 \Bigr | \notag\\
& \quad \leq 2 \bigl | \EE_{\PP_{\btheta}}  (X_j /\sqrt{\sigma_j} )   \bigr  | \cdot \bigl | \EE_{\PP_{\btheta}}  (X_j /\sqrt{\sigma_j} ) - \EE_{\PP_{\btheta}} q_j (Y, \bX) \bigr  |   + \bigl | \EE_{\PP_{\btheta}}  (X_j /\sqrt{\sigma_j} )   - \EE_{\PP_{\btheta}} q_j (Y, \bX) \bigr  | ^2 .
\$
Since $ X_j /\sqrt{\sigma_j} - q_j (Y, \bX)  = X_j /\sqrt{\sigma_j} \cdot \ind\{ |X_j /\sqrt{\sigma_j}| > R \cdot \sqrt{\log d} \}$, by Cauchy-Schwarz inequality we have 
\#\label{eq:trunc1233}
\bigl |  \EE _{\PP_{\btheta}} (X_j /\sqrt{\sigma_j})  - \EE_{\PP_{\btheta}}  q_j (Y, \bX) \bigr | ^2 \leq \EE _{\PP_{\btheta} }( X_j^2  / \sigma_j) \cdot \PP_{\btheta}  \bigl (   |X_j /\sqrt{\sigma_j}| > R \cdot \sqrt{\log d}  \bigr ).
\#
Since $\| \bmu_0\|_{\infty} \vee \| \bmu_1 \|_{\infty} \leq C_0$ and $\{ X_j /\sqrt{\sigma_j} \}_{i=1}^d  $ are  sub-Gaussian random variables, for any $t >0$, there exists a constant $C_1$ such that 
\#\label{eq:trunc1234}
\PP_{\btheta} \bigl( |X_j /\sqrt{\sigma_j}| > t \bigr ) \leq 2 \exp( - C_1 t^2 ).
\#
Thus setting $t = R \cdot \sqrt{\log d }$ for some  sufficiently    large  $R$, by \eqref{eq:trunc1233} and \eqref{eq:trunc1234} we obtain
\$
\bigl |  \EE _{\PP_{\btheta}} (X_j /\sqrt{\sigma_j})  - \EE_{\PP_{\btheta}}  q_j (Y, \bX) \bigr | \leq C_2 d^{-1} 
\$
for some constant $C_2$.   Thus we have 
\#\label{eq:trunc111}
\Bigl |\bigl  [\EE_{\PP_{\btheta}} q_j (Y, \bX)\bigr  ] ^2 -  \bigl[ \EE_{\PP_{\btheta}}  (X_j /\sqrt{\sigma_j} )   \bigr ] ^2 \Bigr | \leq  2 C_0 \cdot C_2 d^{-1} + C_2^2 d^{-2} \leq  1/16  \cdot ( \Delta \bmu)_j^2 / \sigma_j. 
\#
In addition, since $ X_j ^2 / \sigma_j - 1 - \tilde q_j (Y, \bX)  = (X_j ^2 / \sigma_j - 1) \cdot \ind \{ | X_j /\sqrt{\sigma_j}| > R \cdot \sqrt{\log d}\}$, for $\tilde q_j$ defined  in \eqref{eq::query12}, we similarly we obtain
\#\label{eq:trunc112}
\bigl | \EE_{\PP_{\btheta}} \tilde{q}_j (Y, \bX)   -  \EE _{\PP_{\btheta} } ( X_j^2 / \sigma_j - 1 ) \bigr | \leq 1/16  \cdot ( \Delta \bmu)_j^2 / \sigma_j. 
\#
Combining \eqref{eq:trunc111} and \eqref{eq:trunc112} we have
\$ 
\EE_{\PP_{\btheta}} \tilde{q}_j (Y, \bX) - \left  [\EE_{\PP_{\btheta}} q_j (Y, \bX)\right ] ^2   \geq   1/8 \cdot ( \Delta \bmu)_j^2 / \sigma_j~~\text{for all}~~ j\in [d].
\$
 Taking supremum over $j \in [d]$, we have 
\#\label{eq::expected_query3}
  \sup_{j \in [d]}      \left \{ \EE_{\PP_{\btheta}} \tilde{q}_j (Y, \bX) - \left [\EE_{\PP_{\btheta}} q_j (Y, \bX)\right ] ^2 \right\} \geq  1/8 \cdot   \sup_{j \in [d] } \left[ ( \Delta \bmu)_j^2 / \sigma_j \right].
\#
Note that the test function $\overline{\phi}$ involves $4d$ queries functions.
Thus, for any $\btheta  \in \cG_0 (\bSigma) \cup \cG_1(\bSigma; \gamma_n)$, under $\PP_{\btheta} $ the tolerance parameters for $q_j$ and $\tilde{q}_j$ are given by
\#\label{eq:tol_parameter_upper}
\tau_{q_j}  \leq R \sqrt{\log d} \cdot \sqrt{[ \log(4d/\xi)] / n}   ,~~\tau_{\tilde {q}_j} \leq R^2 \log d \cdot   \sqrt{[ \log(4d/\xi)] / n},~\text{for~all~}j \in [d].
\#
Under the assumption  that $${ \sup_{j\in [d]}} ( \Delta \bmu) _j ^2/ \sigma_{j}  = \Omega\left [  \log ^2d \cdot  \log ( d / \xi)/(\alpha^2 n) \wedge \log d \cdot \sqrt{\log( d/\xi)/ n} \right ],$$ we have
\#\label{eq::expect31}
&  \tau_{q_{j} }  \vee  \tau_{ \tilde q_{j} }  \leq   R^2 \log d \cdot   \sqrt{[ \log(4d/\xi)] / n} \notag \\
&\quad\leq  (  1/ C   )\cdot   \left \{   \sup_{j \in [d]}  [  ( \Delta \bmu) _j ^2  / \sigma_j ] \vee \alpha  \cdot  \sup_{j \in [d]}  | ( \Delta \bmu) _j / \sqrt{\sigma_j}  | \right\},
\#
where the absolute constant $C $ is the same as in \eqref{eq::test4}. 
Note that we denote $  R^2 \log d \cdot  \sqrt{\log (4 d / \xi)/n}$ by $\overline\tau _1 $.  Hence by \eqref{eq::expect31}, for any $\btheta \in \cG_0(\bSigma)$, the type-I error of $\overline{\phi}_1$  is bounded by
\$
&\overline{\mathbb{P}}_{\btheta} \left [  \sup_{j \in [d]}  ( Z_{\tilde q_j }   - Z_{  q_j}^2 ) \geq   C  \overline{   \tau} _1\right]  \notag\\
& \quad= \overline{\mathbb{P}}_{\btheta}\left  (  {\textstyle \bigcup_{j \in [d] }} \left \{  ( Z_{\tilde q_j }   - Z_{ q_j}^2 ) -  \left\{ \EE_{\PP_{\btheta}} \tilde{q}_j (Y, \bX)  - [\EE_{\PP_{\btheta}} q_j (Y, \bX)]^2 \right \}   \geq   C \overline{ \tau}_1 \right\} \right ) \notag\\
&\quad \leq \overline{\mathbb{P}}_{\btheta}\left  ( {\textstyle  \bigcup_{j \in [d] }  } \left \{   Z_{\tilde q_j }    - \ \EE_{\PP_{\btheta}} \tilde{q}_j (Y, \bX)   \geq   \overline{ \tau} _1 \right\} \right )+ \overline{\mathbb{P}}_{\btheta}\left (  {\textstyle \bigcup_{j \in [d] }} \left \{   Z_{  q_j }  ^2   - [ \EE_{\PP_{\btheta}}  q_j (Y, \bX)  ]^2 \geq   (C-1)   \overline{ \tau}_1 \right\} \right ).  \$
For the first term, we have
\#\label{eq::sub_oracle_ineq1}
&\overline{\mathbb{P}}_{\btheta}\left( {\textstyle \bigcup_{j \in [d] }} \left\{   Z_{\tilde q_j }    - \ \EE_{\PP_{\btheta}} \tilde{q}_j (Y, \bX)   \geq    \overline{ \tau}_1 \right \} \right ) \notag \\
&\quad \leq  \overline{\mathbb{P}}_{\btheta}\left({\textstyle \bigcup_{j \in [d] }} \left \{ \left |Z_{\tilde q_j}-\mathbb{E}_{\mathbb{P}_{\btheta}}\tilde q_j(Y, \bX)  \right| \geq \tau_{\tilde q_j}\right \} \right) \leq \xi.
\#
Note that under the null hypothesis $\btheta \in \cG_0(\bSigma)$, we have $ \EE_{\PP_{\btheta}}  q_j (Y, \bX)   = \mu_{0, j}/\sqrt{\sigma_j}$. Under the assumption that $\| \bmu_0 \|_{\infty} \vee \| \bmu_1 \|_{\infty} \leq C_0$,  when $n$ is sufficiently large such that 
\$
 \overline \tau _1\leq 3 (C-1)^{-1}  C_0 /\sqrt{\sigma_j}, 
 \$ by  $ Z_{  q_j }  ^2   - [ \EE_{\PP_{\btheta}}  q_j (Y, \bX)  ]^2 \geq    (C-1)  \overline{ \tau}_1 $ we have 
\#\label{eq::intermidiate}
| Z_{q_j} -   \EE_{\PP_{\btheta}}  q_j (Y, \bX)| \geq  (C- 1)\overline\tau_1  \cdot \sqrt{\sigma_{j} }/ (3 C_0).
\# 
Thus we can set absolute constant $ C$ sufficiently large such that 
 $
 | Z_{q_j} -   \EE_{\PP_{\btheta}}  q_j (Y, \bX)|  \geq  \overline \tau_1 .
 $
Thus by \eqref{eq::intermidiate} we have
\#\label{eq::sub_oracle_ineq2}
&\overline{\mathbb{P}}_{\btheta}\Bigl  (  {\textstyle \bigcup_{j \in [d] }} \left\{   Z_{  q_j }  ^2   - [ \EE_{\PP_{\btheta}}  q_j (Y, \bX)  ]^2 \geq  (C-1) \overline{ \tau} _1 \right  \} \Bigr )\notag \\
&\quad\leq \overline{\mathbb{P}}_{\btheta}\left({\textstyle \bigcup_{j \in [d] }} \left  \{ \left |Z_{  q_j}-\mathbb{E}_{\mathbb{P}_{\btheta}} q_j(Y, \bX) \right | \geq  \tau_{  q_j}\right\} \right) \leq \xi.
\#
Combining \eqref{eq::sub_oracle_ineq1} and \eqref{eq::sub_oracle_ineq2}, we can bound the type-I error of $\overline{\phi}_1$ by $2\xi$.
For the type-I error of $\overline{\phi}_2$, 
we define $\bZ= ( 2 Y -1) \cdot   \bX$. Under the data-generating model defined in  \eqref{eq:model_1} and \eqref{eq:model_2},  the distribution of $\bZ$ is given by
\$
\bZ \sim \frac{1+ \alpha}{ 4  }   \Gaussian(-\bmu_0, \bSigma) +\frac{  1+ \alpha}{4} \Gaussian(\bmu_1, \bSigma) +\frac{ 1-\alpha}{4}   \Gaussian( \bmu_0, \bSigma) + \frac{ 1- \alpha}{4} \Gaussian(- \bmu_1, \bSigma).
\$
Then by definition, for all $\btheta \in \cG_0(\bSigma)$,  we have 
\#\label{eq::null3}
\EE_{\prob_{\btheta} } [ \vb ^\top \text{diag}(\bSigma)^{-1/2} \bZ]   = 0, ~\text{for~all}~\vb \in \cB_2(1).
\#  
In addition, for any $\btheta \in \cG_1(\bSigma; \gamma_n)$, by the distribution of $\bZ$,  for all $\vb\in \cB_2(1)$, we have 
\#\label{eq::alt3}
  \EE_{\prob_{\btheta} } [\vb^\top   \text{diag}(\bSigma)^{-1/2} \bZ]   =  \alpha  /2\cdot \vb ^\top   \text{diag}(\bSigma)^{-1/2}\Delta \bmu.
\#
Moreover, by definition we have
\$\vb ^\top  \text{diag}(\bSigma)^{-1/2} \bZ - \overline{q}_\vb (Y, \bX)  = \vb ^\top  \text{diag}(\bSigma)^{-1/2} \bZ \cdot \ind \bigl  \{ |\vb ^\top  \text{diag}(\bSigma)^{-1/2} \bZ| \leq R \sqrt{\log d} \bigr \}  .
\$
By setting the constant $R$ sufficiently large, for any  for any $\btheta \in \cG_0(\bSigma) \cup\cG_1(\bSigma; \gamma_n)$, we have
\$
\bigl | \EE_{\prob_{\btheta}} \overline{q}_\vb (Y, \bX) - \EE _{\prob_{\btheta}} ( \vb ^\top  \text{diag}(\bSigma)^{-1/2} \bZ )  \bigr | \leq \alpha/4\cdot  | ( \Delta \bmu) _j / \sqrt{\sigma_j}  |.
\$
Combining \eqref{eq::null3} and \eqref{eq::alt3} we obtain that 
\$
&\EE_{\prob_{\btheta}} \overline{q}_\vb (Y, \bX)  \leq  \alpha/4\cdot  | ( \Delta \bmu) _j / \sqrt{\sigma_j}  |~~\text{for all}~~\btheta \in \cG_0(\bSigma);\\
 &\EE_{\prob_{\btheta}} \overline{q}_\vb (Y, \bX)   \geq \alpha/4\cdot  | ( \Delta \bmu) _j / \sqrt{\sigma_j}  |~~\text{for all}~~\btheta \in \cG_1( \bSigma, \gamma_n).
\$
Thus, taking the supremum over $\cB_2(1)$ yields
\#\label{eq::sup_diff}
  \sup_{\vb \in \cB_2(1)}  \EE_{\prob_{\theta}} \overline{q}_\vb (Y, \bX)   \geq  \alpha/4\cdot \sup_{j \in [d]}  | ( \Delta \bmu) _j / \sqrt{\sigma_j}  |~~\text{for all}~~\btheta \in \cG_1( \bSigma, \gamma_n).
\#
In addition, since we have $4d$ queries, by Definition \ref{def::oracle},  the tolerance parameters for $\overline{q}_{\vb}$'s   are bouded by
\$
\tau_{\overline{q}_{\vb}} \leq  R \sqrt{\log d}\cdot  \sqrt{[ \log(4d/\xi)] / n},~\text{for~all~} \vb \in \cB_2(1).
\$
Note that we denote $\overline{\tau}_2 = R \sqrt{\log d}\cdot  \sqrt{[ \log(4d/\xi)] / n}$. 
Similar to  \eqref{eq::expect31},   we have 
\#\label{eq::expect3}
\tau_{\overline{q}_{\vb}}  \leq \overline{\tau} _1 \leq       (1/ C)\cdot \left \{ \sup_{j \in [d]} [  ( \Delta \bmu) _j ^2  / \sigma_j ] \vee \sup_{j \in [d]} \alpha | ( \Delta \bmu) _j / \sqrt{\sigma_j}  | \right\}.
\#
Hence by \eqref{eq::expect3}, for any $\btheta \in \cG_0(\bSigma)$, the type-I error of $\overline{\phi}_2$  is bounded by
\#\label{eq::bound_prob_oracle1}
&\overline{\mathbb{P}}_{\btheta} \left( \sup_{\vb \in \cB_2(1)} Z_{\overline{q} _{\vb}}   \geq2 \overline{\tau}_1 \right)   = \overline{\mathbb{P}}_{\btheta}\left({\textstyle \bigcup_{\vb \in \cB_2(1)}}  \left\{Z_{\overline{q} _{\vb}} -  \mathbb{E}_{\mathbb{P}_{\btheta }} [\overline{q}_{\vb}(Y, \bX)]  >    \overline{\tau}_1  \right\} \right ) \notag\\
&  \leq \overline{\mathbb{P}}_{\btheta}\left({\textstyle \bigcup_{\vb \in \cB_2(1)} }  \left \{\left  |Z_{\overline{q}_{\vb}}-\mathbb{E}_{\mathbb{P}_{\btheta}}[\overline{q}_{\vb} (Y,  \bX)]  \right| \geq  \tau_{\overline{q}_{\vb}} \right\} \right) \leq \xi.
\#
Combining \eqref{eq::sub_oracle_ineq1},  \eqref{eq::sub_oracle_ineq2}, and  \eqref{eq::bound_prob_oracle1}, we have 
\$
\overline{\PP}_{\btheta} ( \overline{\phi} = 1) \leq 3\xi, ~\text{for~all~} \btheta \in \cG_0(\bSigma).
\$
\paragraph{Type-II error.}  
Now we consider $\btheta \in \parSpace_1(\bSigma; \gamma_n)$. Note that $\overline \phi = 0$ if $\overline\phi_1 = 0$ and $\overline \phi_2 = 0$. Thus, for any $\btheta \in \parSpace_1(\bSigma; \gamma_n)$, we have
\begin{equation*}
\overline \prob_{\btheta} ( \overline \phi = 0) =  \overline\prob_{\btheta} (\overline\phi_1 = 0 \cap \overline\phi_2 = 0)  \leq \overline\prob_{\btheta}( \overline\phi_1 = 0 ) \wedge \overline\prob_{\btheta} ( \overline\phi_2 = 0).
\end{equation*}
Recall that we denote $\Delta \bmu = \bmu_1 - \bmu_0$. Similar to the proof of Theorem  \ref{thm:info_upper_bound}, we  consider two cases of  the condition  \$    \sup_{j\in [d]} ( \Delta \bmu) _j ^2/ \sigma_{j}  = \Omega\left [  \log ( d / \xi)/(\alpha^2 \cdot n) \wedge  \sqrt{\log( d/\xi)/ n} \right]. \$

\paragraph{Case (i).} We show that the  type-II error of $\overline \phi_1$ is negligible  under the assumption that   \${\sup_{j\in [d]}  } ( \Delta \bmu) _j ^2/ \sigma_{j}   = \Omega \left  [   \sqrt{\log( d/\xi)/ n} \right].\$
Let $j^* = \argmax _{j\in[d] }  ( \Delta \bmu )_j^2 /  \sigma_{j}$.~Then by \eqref{eq::expect31}, when  we have 
\#\label{eq::diff_mean2}
1 +C \overline{\tau} \leq  (\Delta \bmu)_{j^*}^2 / \sigma_{j^*} + 1 - C \overline{\tau} = \EE_{\PP_{\btheta} } \tilde q_{j^*} ( Y, \bX) - [\EE_{\PP_{\btheta}}   q_{j^*} ( Y, \bX) ]^2   - C\overline{\tau}. 
\#
Thus combining  \eqref{eq:trunc111}, \eqref{eq:trunc112}, and  \eqref{eq::diff_mean2}, we have
\#\label{eq::bound_prob_oracle2}
& \overline{\mathbb{P}}_{\btheta} \Bigl [  \sup_{j \in [d]}  ( Z_{q_j }   - Z_{\tilde q_j}^2 ) <  C \overline{   \tau}_1 \Bigr]  \notag\\
& \quad \leq  \overline{\mathbb{P}}_{\btheta} \left  \{    Z_{\tilde q_j^* }   - Z_{  q_j}^2    <  \EE_{\PP_{\btheta} } \tilde q_{j^*} ( Y, \bX) - [\EE_{\PP_{\btheta}}   q_{j^*} ( Y, \bX) ]^2   - C\overline{\tau}_1  \right\} \notag\\
&\quad \leq\overline{\mathbb{P}}_{\btheta}  \left   [  \EE_{\PP_{\btheta} } \tilde q_{j^*} ( Y, \bX)- Z_{\tilde q_{j^*}}  >  \overline{\tau}_1 \right ]  +  \overline{\mathbb{P}}_{\btheta}   \left \{    [\EE_{\PP_{\btheta}}   q_{j^*} ( Y, \bX) ]^2 -  Z_{q _j ^*} ^2>    (C-1) \overline{\tau} _1\right \} .
\#
Moreover,  by \eqref{eq::expect31} the first term on the right-hand side of \eqref{eq::bound_prob_oracle2} can be  further bounded by
\#\label{eq::bound_prob_oracle22}
& \overline{\mathbb{P}}_{\btheta}    \left [  \EE_{\PP_{\btheta} } \tilde q_{j^*} ( Y, \bX)- Z_{\tilde q_{j^*}}  > \overline{\tau}_1 \right ]  \notag\\
&\quad \leq  \overline{\mathbb{P}}_{\btheta}   \left \{ \EE_{\PP_{\btheta} } \tilde q_{j^*} ( Y, \bX)- Z_{\tilde q_{j^*}}  \geq   \tau_{\tilde q_{j^*}} \right \} \notag\\
&\quad  \leq  \overline{\mathbb{P}}_{\btheta}    \left(\textstyle{\bigcup_{j\in[d]}} \left \{ |Z_{ \tilde q_{j}}-\EE_{\PP_{\btheta} } \tilde q_j(Y, \bX)] | \geq  \tau_{\tilde q_j} \right\} \right) \leq \xi.
\#
Similarly, for the second term on the right-hand side of \eqref{eq::bound_prob_oracle2}, by   \eqref{eq::expect31} and \eqref{eq::intermidiate}we have 
\#\label{eq::bound_prob_oracle23}
&\overline{\mathbb{P}}_{\btheta}   \left \{    [\EE_{\PP_{\btheta}}   q_{j^*} ( Y, \bX) ]^2 -  Z_{q _j ^*} ^2>   (C- 1) \overline{\tau} _1 \right\}  \notag\\
&\quad  \leq  \overline{\mathbb{P}}_{\btheta}    \left(\textstyle{\bigcup_{j\in[d]}} \left \{ \left |Z_{q_{j}}-\EE_{\PP_{\btheta} } \tilde q_j(Y, \bX)] \right| \geq \tau_{q_j}\right\} \right) \leq \xi.
\#
Therefore, combining \eqref{eq::bound_prob_oracle22} and \eqref{eq::bound_prob_oracle23}, we conclude that the type-II error of $\overline \phi_2$ is no more than $2 \xi$.

\paragraph{Case (ii).} Now we assume study the type-II error of $\overline \phi_2$ under the assumption that  \$ \sup_{j\in [d]}( \Delta \bmu) _j ^2/ \sigma_{j}    = \Omega \left  [ \log ( d / \xi)/(\alpha^2 \cdot n) \right] .\$
Let $j^* = \argmax _{j\in[d] }  ( \Delta \bmu )_j^2 /  \sigma_{j}$ and $\vb^* = \argmax _{\vb \in \cB_2(1) }\EE_{\prob_{\theta}} \overline{q}_\vb (Y, \bX)$. Then by  \eqref{eq::expect31}  and \eqref{eq::sup_diff}, when  $C > 4$ we have
\#\label{eq::compute_mean222}
2 \overline \tau_2 \leq  \alpha/2\cdot {\sup_{j \in [d]} } | ( \Delta \bmu) _j / \sqrt{\sigma_j}  | -2 \overline \tau_2 = \EE_{\prob_{\theta}} \overline{q}_{\vb ^*} (Y, \bX) -2  \overline \tau_2.
\# 
Then by \eqref{eq::expect3} and \eqref{eq::compute_mean222} the type-II error of $\overline \phi_2$ is bounded by
\#\label{eq::bound_type22}	
&\overline{\mathbb{P}}_{\btheta}\left  (   \sup_{\vb \in \cB_2(1)}  Z_{\overline{q} _{\vb}}   <  2 \overline{\tau}_2 \right )  \leq  \overline{\mathbb{P}}_{\btheta} \left  [ \sup_{\vb \in \cB_2(1)} Z_{\overline{q} _{\vb}}  < \EE_{\prob_{\theta}} \overline{q}_{\vb ^*} (Y, \bX) - 2 \overline \tau _2\right]\notag \\
&\quad  \leq \overline{\mathbb{P}}_{\btheta}  \left [ Z_{\overline{q} _{\vb^*}}  < \EE_{\prob_{\theta}} \overline{q}_{\vb ^*} (Y, \bX) -2  \overline \tau _2\right ] \notag \\
&\quad  \leq \overline{\mathbb{P}}_{\btheta}\left({\textstyle \bigcup_{\vb \in \cB_2(1)} } \left \{ \left |Z_{\overline{q}_{\vb}}-\mathbb{E}_{\mathbb{P}_{\btheta}}[\overline{q}_{\vb} (Y,  \bX)] \right  | \geq  \tau_{\overline{q}_{\vb}}\right\} \right) \leq \xi.
\#
Thus by \eqref{eq::bound_type22},   the type-II error of $\overline {\phi}_2$ is no more than $\xi$.  Then together with Case (i), we have $\overline \PP_{\btheta} ( \overline{\phi}) \leq 2\xi$ for all $\btheta \in \cG_1(\bSigma; \gamma_n)$. Therefore the total risk of $\overline{\phi}$ is bounded by 
\$
\overline{R}_n (\overline{\phi}) =  \sup_{\btheta \in \cG_0(\bSigma) } \overline{\PP}_{\btheta} ( \overline{\phi} = 1) +   \sup_{\btheta \in \cG_1(\bSigma;\gamma_n)}  \overline{\PP}_{\btheta} ( \overline{\phi} = 0)\leq 5 \xi.
\$

\section{Proofs for Technical Lemmas}
In this section, we prove the technical lemmas which appear in the proofs of the main results.
\subsection{Proof of Lemma \ref{lem::h_function} } \label{proof::lem::h_function}
Under $\PP_{\bf 0}$, $\bX$ and $Y$ are independent with $\bX \sim \Gaussian ( \bf 0, \Ib)$ and $Y $ is uniform over~$\{ 0 , 1\}$.
We denote by $f ( \xb; \bmu )$ the density of $\Gaussian ( \bmu, \Ib)$ and by $p_{\bf 0}(y, \xb)$ the density of $\PP_{\bf 0}$. Then for any~$y \in \{0, 1\}$~and~$\xb \in \RR^d$, we have
$ 
p_{\bf 0} ( y , \xb) = 1/2 \cdot f (\xb ;\bf 0).
$
In addition, for any $\vb \in \cH(s)$, we denote the density of $\PP_{\vb }$~by~$p_{\vb} ( y, \xb)$.  By the definition of the statistical model, we have
\$
p_{\vb} (1, \xb) & =   ( 1+ \alpha) /4\cdot f(  \xb;     \vb /2 )  + ( 1-\alpha) /4\cdot f (  \xb;  -   \vb /2 ),   \\
p_{\vb} (0, \xb) & = ( 1-\alpha) /4 \cdot f(  \xb;     \vb /2 ) + ( 1+ \alpha) /4\cdot f (  \xb;  -   \vb /2 ).
\$
Thus for any $y \in \{0,1\}$ and $\xb \in \RR^d$, we have
\#\label{eq::likelihood_ratio}
\frac{\dd \PP_{\vb}} {\dd \PP_{\bf 0}} ( y , \xb) = \frac{1  }{2} \cdot  \left[ \frac{ f( \xb;    \vb /2 )}{ f(\xb; {\bf 0})} + \frac{ f (\xb; -   \vb /2 ) }{ f(\xb; {\bf 0})} \right]   + \frac{ \alpha ( 2 y -1) }{2}  \cdot  \left[\frac{ f( \xb;    \vb /2 )}{ f(\xb; {\bf 0})} - \frac{ f (\xb; -   \vb /2 ) }{ f(\xb; {\bf 0})} \right].
\#
Note that by definition, for any $\bmu \in \RR^d$, we have 
\$
g(\xb; \bmu ) : = f(\xb; \bmu ) / f(\xb; {\bf 0} ) =\exp(   \bmu ^\top \xb  - 1/2 \cdot  \| \bmu \|_2^2 ) .
\$
Thus \eqref{eq::likelihood_ratio} is reduced to 
\#\label{eq::likelihood_ratio2}
\frac{\dd \PP_{\vb}} {\dd \PP_{\bf 0}} ( y , \xb) = \left [ g( \xb, \vb /2 ) + g( \xb;  -\vb/2)\right] /2 + \alpha (2 y -1) \cdot \left  [ g( \xb, \vb /2 ) - g( \xb;  -\vb/2)\right] /2.
\#
For any $\vb_1, \vb_2 \in \cH(s)$,  by \eqref{eq::likelihood_ratio2} we have 
\#\label{eq::likelihood_ratio3}
& \EE_{\PP_{\bm 0}} \left[ \frac{\ud \PP_{\vb_1}}{\ud \PP_{ \bm 0}} \frac{\ud \PP_{\vb_2}}{\ud \PP_{\bm 0}} ( Y , \bX) \right] \notag \\
& \quad =  \EE_{\PP_{\bm 0}}    \left \{ \left [ g( \bX, \vb_1  /2 ) + g( \bX ;  -\vb_1/2)\right] \cdot \left  [ g( \bX, \vb_2  /2 ) + g( \bX ;  -\vb_2/2)\right] /4  \right \}\notag \\
&\quad\quad \quad + \alpha^2 \cdot \EE_{\PP_{\bm 0}} \left  \{ \left  [ g( \bX, \vb_1  /2 ) - g( \bX ;  -\vb_1/2)\right] \cdot \left   [ g( \bX, \vb_2  /2 ) - g( \bX ;  -\vb_2/2)\right ] /4 \right \}, 
\#
where we use the independence of $Y$ and $\bX$ under $\PP_{\bf 0} $.	
In what follows, we calculate the two terms on the right-hand side of \eqref{eq::likelihood_ratio3}, respectively.
Let $\eta_1$ and $ \eta_2$ be two independent Rademacher random variables over $\{ -1, 1\}$. Then for $\ell \in \{1, 2\}$, we have 
\#
\left [g( \bX, \vb_{\ell}  /2 ) + g( \bX ;  -\vb_{\ell} /2)\right ]/2 &= \EE_{\eta _{\ell}} \left [ g( \bX, \eta _{\ell} \vb _{\ell} /2)\right] , \label{eq::use_eta1}\\
\left [g( \bX, \vb_{\ell}  /2 ) -  g( \bX ;  -\vb_{\ell} /2)\right ]/2&= \EE_{\eta _{\ell}} \left  [ \eta _{\ell} \cdot g( \bX, \eta _{\ell}\vb _{\ell} /2)\right].\label{eq::use_eta2}
\#
Then by 	\eqref{eq::use_eta1} and \eqref{eq::use_eta2} we have 
\#\label{eq::use_mgf}
& \EE_{\PP_{\bm 0}}   \left  \{ \left [ g( \bX, \vb_1  /2 ) + g( \bX ;  -\vb_1/2) \right] \cdot  \left [ g( \bX, \vb_2  /2 ) + g( \bX ;  -\vb_2/2)\right] /4  \right  \} \notag \\
& \quad= \EE_{\PP_{\bf 0}}\EE_{\eta _{1} , \eta_{2}} \left  [ g(\bX ; \eta _{1}  \vb _1/ 2) \cdot g(\bX ; \eta _{2}  \vb _2/ 2) \right] \notag\\
& \quad= \EE_{\eta _{1} , \eta_{2}}\EE_{\PP_{\bf 0}}  \exp \left  [  \bX ^\top (\eta_1  \vb_1 +  \eta_2  \vb_2) /2 -  1/8 \cdot ( \|\vb_1 \|_2^2 + \| \vb_2\|_2^2)  \right].
\#
Using the moment-generating function of $\bX$, by \eqref{eq::use_mgf} we have 
\$
& \EE_{\PP_{\bm 0}}    \left  \{ \left [ g( \bX, \vb_1  /2 ) + g( \bX ;  -\vb_1/2)\right] \cdot \left   [ g( \bX, \vb_2  /2 ) + g( \bX ;  -\vb_2/2)\right] /4 \right  \} \notag \\
& \quad= \EE_{\eta _{1} , \eta_{2}} \left [ \exp ( 1/2 \cdot \eta_1 \eta _2 \cdot \vb_1 ^\top \vb_2 ) = \cosh( 1/2 \cdot \la \vb_1 , \vb_2 \ra )\right].
\$
Similarly, for \eqref{eq::use_eta2} we have
\$
& \EE_{\PP_{\bm 0}} \left \{ \left [ g( \bX, \vb_1  /2 ) - g( \bX ;  -\vb_1/2)] \cdot  [ g( \bX, \vb_2  /2 ) - g( \bX ;  -\vb_2/2)\right ] /4 \right\} \\
& \quad= \EE_{\PP_{\bf 0}}\EE_{\eta _{1} , \eta_{2}} \left [\eta_1  \eta _2 \cdot  g(\bX ; \eta _{1}  \vb _1/ 2) \cdot g(\bX ; \eta _{2}  \vb _2/ 2)\right ]  \\
& \quad= \EE_{\eta _{1} , \eta_{2}} \left [\eta _1 \eta_2\cdot \exp ( 1/2 \cdot \eta_1\eta _2 \cdot \vb_1 ^\top \vb_2 ) \right] = \sinh( 1/2 \cdot \la \vb_1 , \vb_2 \ra ).
\$
Thus we conclude the proof of Lemma \ref{lem::h_function}.

\subsection{Proof of Lemma \ref{lem:h_upper_bound}} \label{proof:lem:h_upper_bound}
It is straightforward to verify \eqref{eq:max} holds when $x = 0$. We focus on region $x > 0$. It is then sufficient to prove the result for these two cases below.

\paragraph{Case 1:} We consider the case $v \leq 1/ (2x) \cdot  \log [ \cosh(2x)]$. Then we need to prove
\begin{equation} \label{eq:case_1}
\cosh(x) + v\sinh(x) \leq \cosh(2x).
\end{equation}
Using the bound of $v$, it remains to show the function
\[
f(x) = 1 / (2x) \cdot \log[ \cosh(2x)] \cdot  \sinh(x)  + \cosh(x) - \cosh(2x) \leq 0.
\]
holds for all $x > 0$. It's easy to verify $f(x)$ is monotonically decreasing over $(0,\infty]$ and $\lim_{x \rightarrow 0}f(x) = 0$. We thus finish proving \eqref{eq:case_1}.

\paragraph{Case 2:} We consider the case $v \geq 1 / (2x) \cdot  \log[ \cosh(2x)]$. We would like to show
\begin{equation} \label{eq:case_2}
\cosh(x) + v\sinh(x) \leq \exp( {2vx}).
\end{equation}
Let us define $g(v) := \exp( {2vx} ) - \cosh(x) - v\sinh(x)$. We have that for any $x \geq 0$,
\[
g'(v) = 2x\exp( {2vx}) - \sinh(x) \geq 2x\cosh(2x) - \sinh(x) \geq 0.
\]
Hence, $g(v)$ is a monotonically increasing function. We thus have 
\$
g(v) &\geq g\left\{ 1/ (2x) \cdot \log[ \cosh(2x)]\right\} \\
&  = \cosh(2x) - \cosh(x) - 1/(2x) \cdot \log[ \cosh(2x)] \cdot \sinh(x) = -f(x) \geq 0.
\$
We thus finish proving \eqref{eq:case_2}.

\section{Supporting Lemmas}
In this section we list the supporting lemmas that establish two concentration inequalities for Gaussian random variables.
\begin{lemma}[$\chi^2$-tail bound, \cite{johnstone1994minimax}] \label{lem:chi-square}
	Let $X_1, \ldots, X_n$ be $n$ i.i.d. standard normal random variables. For all $t \in (0,1)$,
	\[
	\prob\left( \left| \frac{1}{n}\sum_{i=1}^n X_i^2 - 1\right| \geq t\right) \leq 2 \exp( {-nt^2/8}).
	\]
\end{lemma}

\begin{lemma}[Gaussian covariance estimation, \cite{vershynin2010introduction}] \label{lem:Gaussian_cov}
	Suppose $\{\bX_i\}_{i=1}^n$ are $n$ i.i.d. Gaussian random vectors in $\real^d $ and $\bX_1 \sim \Gaussian(\bm{0}, \bSigma)$. For every $\epsilon \in (0,1)$, and $t \geq 1$, if $n \geq C(t/\epsilon)^2d$ for some constant $C$, then with probability at least $1 - 2e^{-t^2n}$,
	\[
	\opnorm{\widehat{\bSigma} - \bSigma}{2} \leq \epsilon\opnorm{\bSigma}{2},
	\]
	where $\widehat{\bSigma} := 1/n\cdot\sum_{i=1}^n\bX_i\bX_i^{\top}$.
\end{lemma}